\documentclass[conference]{IEEEtran}
\IEEEoverridecommandlockouts
\AtBeginDocument{%
  \providecommand\BibTeX{{%
    \normalfont B\kern-0.5em{\scshape i\kern-0.25em b}\kern-0.8em\TeX}}}

\usepackage{booktabs}
\usepackage{tabulary}
\usepackage{lscape}
\usepackage{mathtools}
\usepackage{seqsplit}
\usepackage{amsmath,amssymb,amsfonts}
\usepackage{dirtytalk}
\usepackage{enumitem} 
\usepackage{xspace}
\usepackage{xcolor}
\usepackage{multirow}
\usepackage[normalem]{ulem} 
\usepackage[switch]{lineno} 

\usepackage{tabularx}
\usepackage[flushleft]{threeparttable}

\newcommand{\fig}[1]{Fig.~\ref{#1}}

\newcommand{\equationref}[1]{Eq.~\ref{#1}}
\newcommand{\sect}[1]{\S\ref{#1}}
\newcommand{\etal}[0]{{\em et al.~}}
\newcommand{\eg}[0]{{\em e.g.,~}}
\newcommand{\ie}[0]{{\em i.e.,~}}

\newcommand{\modelname}{{CORAL\xspace}}

\newcommand{\projectsite}{{http://bdata.cs.washington.edu/coral/}}
\newcommand{\modelnamelong}{{COde RepresentAtion Learning with weakly-supervised transformers\xspace}}

\definecolor{codecolor}{RGB}{156,218,255}
\definecolor{markdowncolor}{RGB}{248,197,94}
\definecolor{weaksupervisioncolor}{RGB}{212,194,255}
\definecolor{unsupervisedreconstruction}{RGB}{130,206,255}

\newcommand{\IMPORT}{{\texttt{IMPORT}}}
\newcommand{\WRANGLE}{{\texttt{WRANGLE}}}
\newcommand{\EXPLORE}{{\texttt{EXPLORE}}}
\newcommand{\MODEL}{{\texttt{MODEL}}}
\newcommand{\EVALUATE}{{\texttt{EVALUATE}}}

\graphicspath{{./fig/}}

\newcommand{\hide}[1]{}
\newcommand{\xhdr}[1]{\vspace{1.7mm}\noindent{{\bf #1.}}}

\begin{document}

\title{CORAL: COde RepresentAtion Learning with Weakly-Supervised Transformers for Analyzing Data Analysis}

\author{
\IEEEauthorblockN{Ge Zhang\IEEEauthorrefmark{1}\textsuperscript{\textsection},
Mike A. Merrill\IEEEauthorrefmark{2}\textsuperscript{\textsection},
Yang Liu\IEEEauthorrefmark{2},
Jeffrey Heer\IEEEauthorrefmark{2},
Tim Althoff\IEEEauthorrefmark{2}}
\IEEEauthorblockA{\IEEEauthorrefmark{1}Department of Computer Science\\
Peking University, Beijing, China\\
zhangge9194@pku.edu.cn}
\IEEEauthorblockA{\IEEEauthorrefmark{2}Paul G. Allen School of Computer Science and Engineering\\
University of Washington, Seattle, Washington,\\
\{mikeam,yliu0,jheer,althoff\}@cs.washington.edu}
}



\maketitle
\begingroup\renewcommand\thefootnote{\textsection}
\footnotetext{These authors contributed equally to this work.}
\endgroup
\begin{abstract}

Large scale analysis of source code, and in particular scientific source code, holds the promise of better understanding the data science process, identifying 
analytical best practices, and providing insights to the builders of scientific toolkits. 
However, large corpora have remained unanalyzed in depth, as descriptive labels are absent and require expert domain knowledge to generate. We propose a novel weakly supervised transformer-based architecture for computing joint representations of code from both abstract syntax trees and surrounding natural language comments. We then evaluate the model on a new classification task for labeling computational notebook cells as stages in the data analysis process from data import to wrangling, exploration, modeling, and evaluation. We show that our model, leveraging only easily-available weak supervision, achieves a 38\% increase in accuracy over expert-supplied heuristics and outperforms a suite of baselines. Our model enables us to examine a set of 118,000 Jupyter Notebooks to uncover common data analysis patterns. Focusing on notebooks with relationships to academic articles, we conduct the largest ever study of scientific code and find that notebook composition correlates with the citation count of corresponding papers. 
\end{abstract}






\newcommand{\figCoral}{
\begin{figure}
\vspace{-5pt}
    \includegraphics[width=0.99\columnwidth]{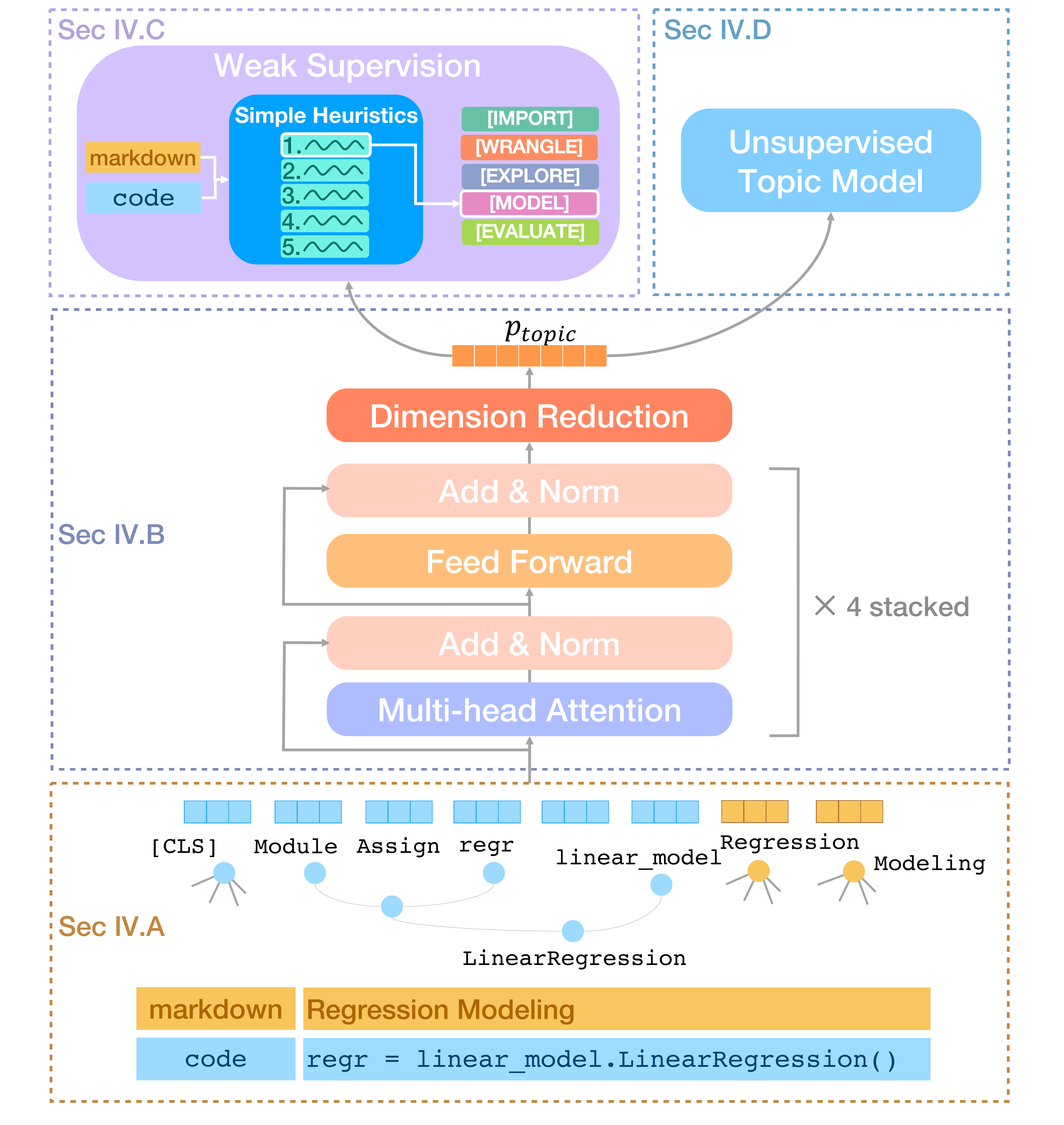}
    \vspace{-10pt}
    \caption{An overview of the architecture of our \modelname{} model, which combines \textcolor{weaksupervisioncolor}{\textbf{weak supervision}} and \textcolor{unsupervisedreconstruction}{\textbf{unsupervised topic modeling}} into a multitask objective. For visual clarity, we only show edges from the AST here. In practice, we also use connections between [CLS] and all the others nodes, and between each AST node and markdown node (see Section~\ref{subsec:input_graphs}).
    }
    \vspace{-15pt}
    \label{fig:coral}
\end{figure}
}

\definecolor{weaksupervisioncolor}{RGB}{212,194,255}
\definecolor{unsupervisedreconstruction}{RGB}{130,206,255}

\newcommand{\figTask}{
\begin{figure}
    \centering
    \includegraphics[width = 0.9\columnwidth]{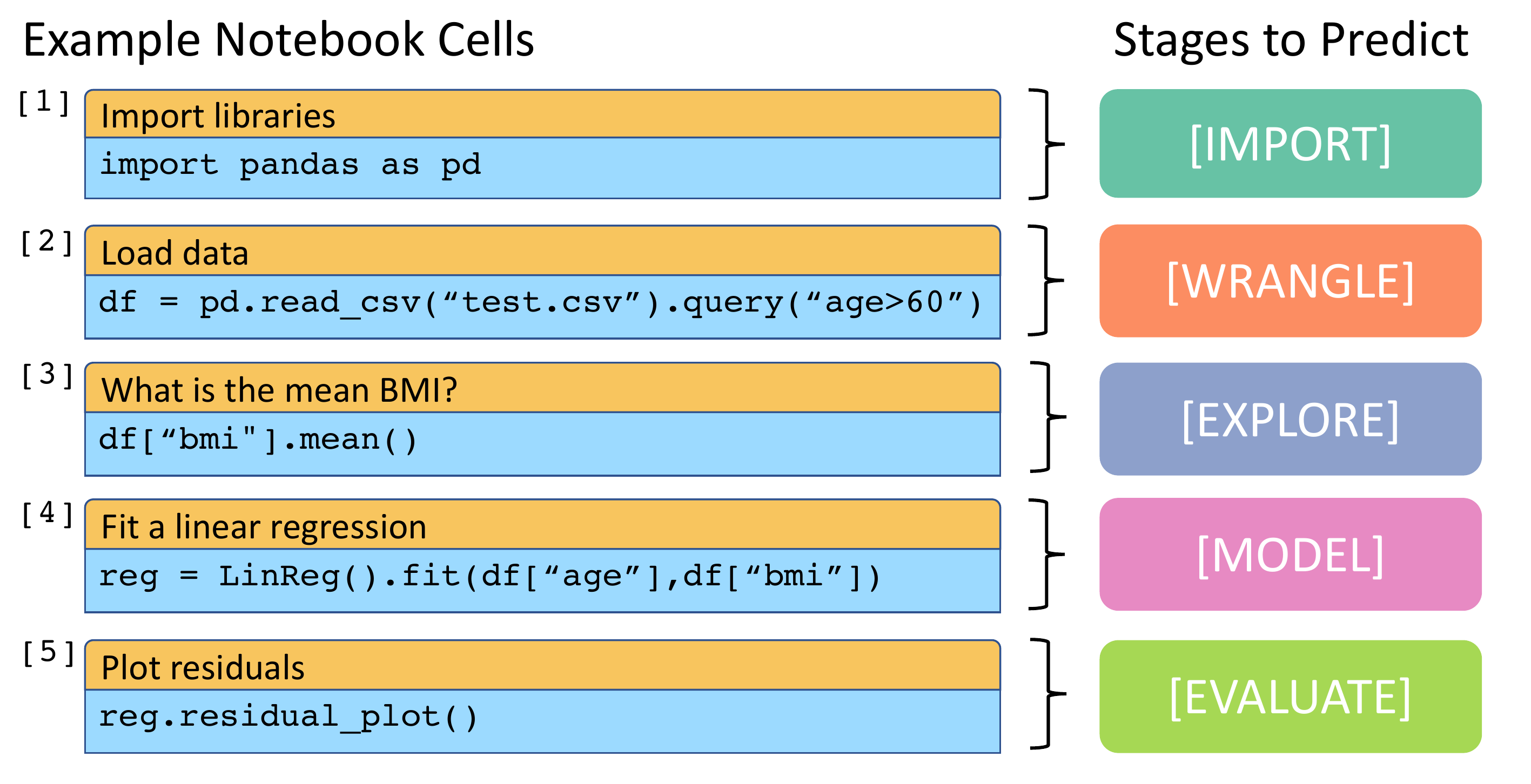}
    \vspace{-12pt}
    \caption{Examples of our proposed task of automatically labeling code snippets and accompanying natural language annotations as stages in the data science process, with code in \textcolor{codecolor}{\textbf{blue}} and markdown in \textcolor{markdowncolor}{\textbf{yellow}}.}
    \vspace{-15pt}
    \label{fig:task_description}
\end{figure}
}

\newcommand{\figBaseline}{
\begin{figure}
    \centering
    \vspace{-10pt}
    \includegraphics[width=0.99\columnwidth]{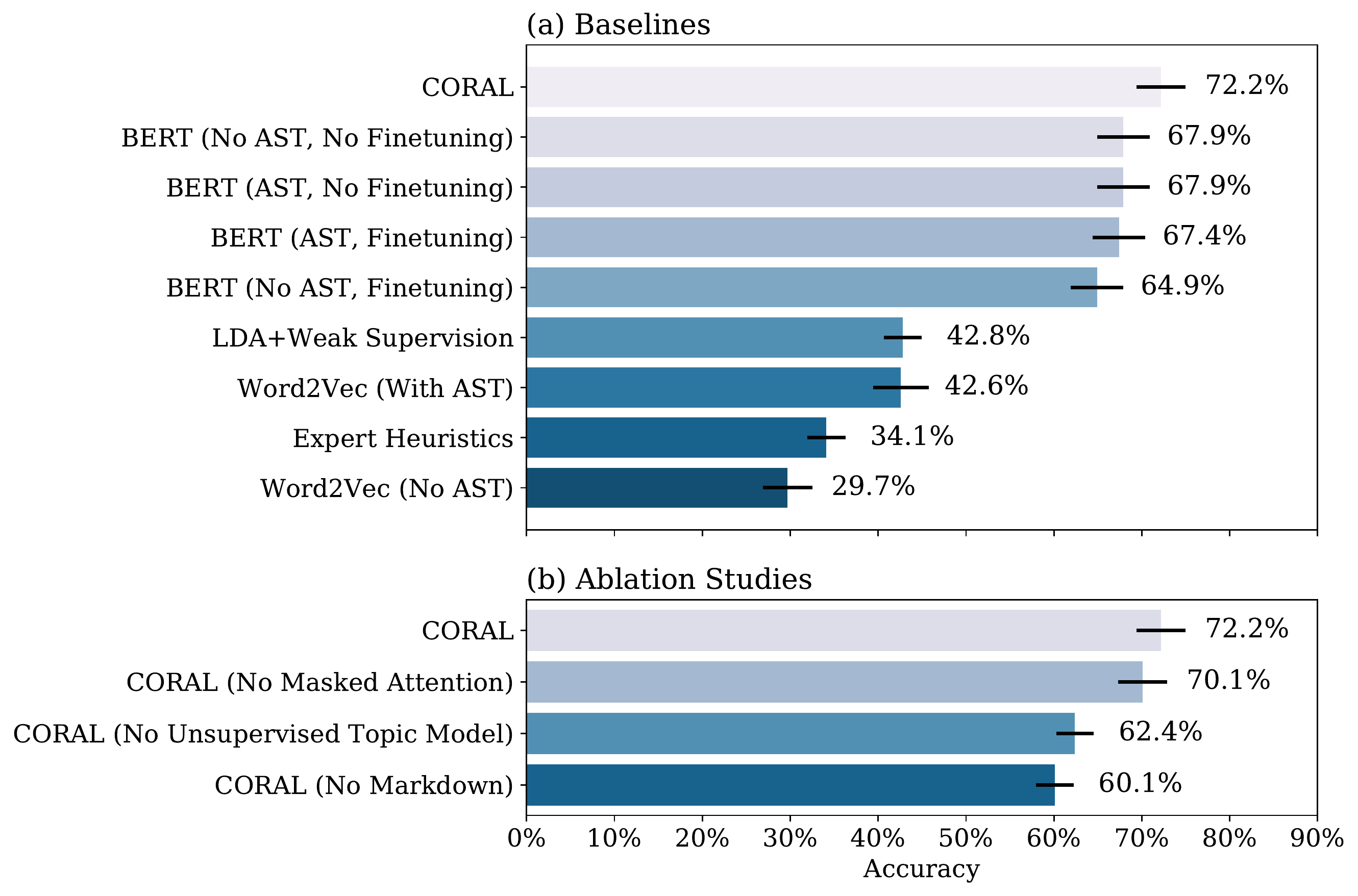}
    \vspace{-20pt}
    \caption{Accuracy on expert-annotated test set for all baselines (a) and ablation studies (b). Performance improves with neural topic models and weak supervision. \modelname\ significantly outperforms all baselines (Wilcoxon signed rank, $p<0.001$) and ablation studies ($p<0.05$)}
    \vspace{-15pt}
    \label{fig:performance_with_baselines}
\end{figure}
}

\newcommand{\figPseudoCode}{
\begin{figure}[ht]
    \centering

    \includegraphics[width=0.5\columnwidth]{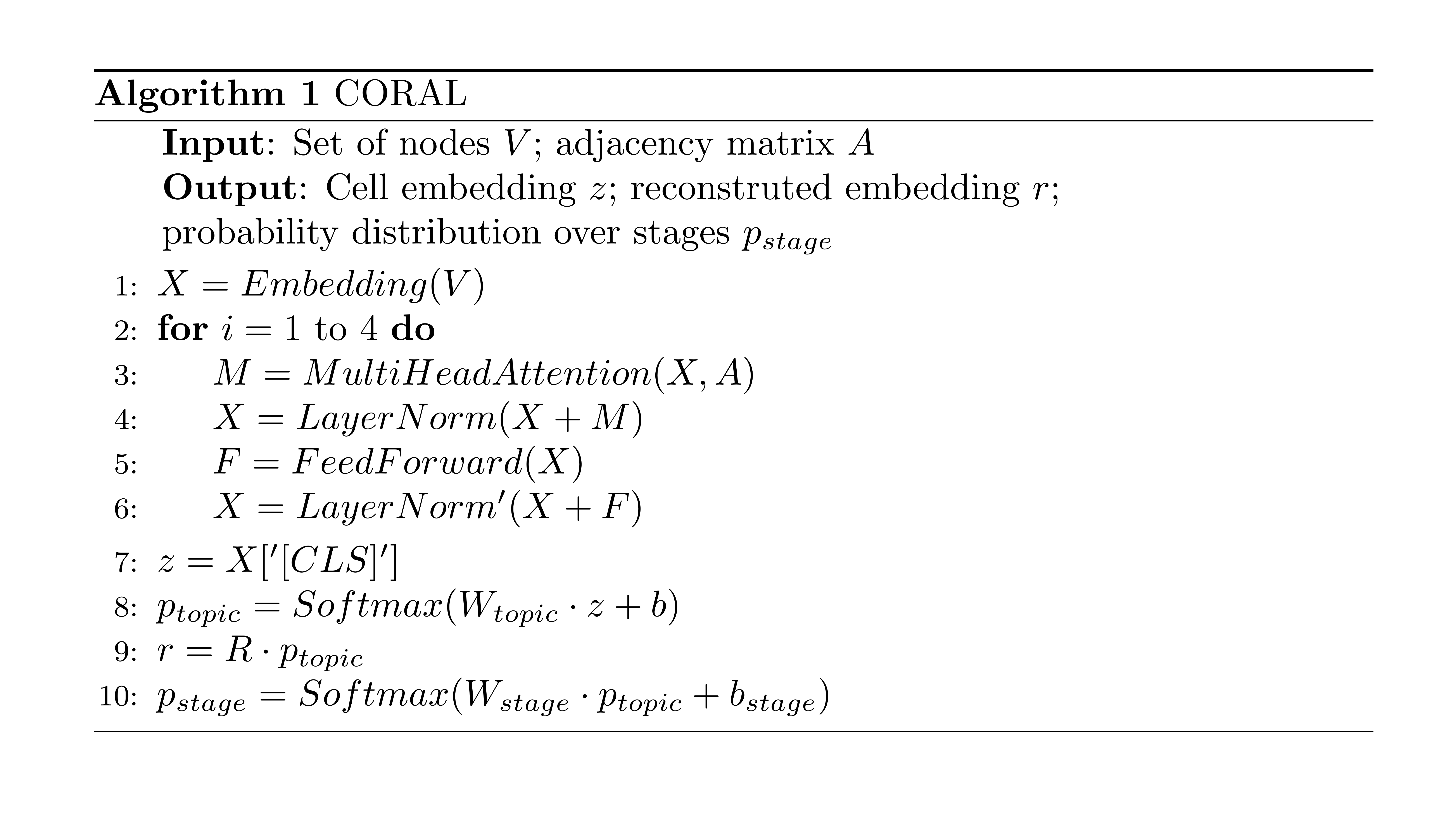}
    \vspace{-20pt}
    \caption{CORAL Algorithm.}
    \vspace{-15pt}
    \label{fig:pseudo_code}
\end{figure}
}

\newcommand{\tableMaxSeq}{
\begin{table}[]
    \centering
    \begin{threeparttable}
    \vspace{-10pt}
    \caption{Impact of Max Sequence Length on \modelname~}     
        \begin{tabularx}{0.99\columnwidth}{@{\extracolsep{\fill}}l c c c}
        \hline
        Model & \multicolumn{3}{c}{Max Sequence Length} \\
         & 80 & 120 & 160 \\
        \hline
        CORAL & 59.9 & 64.2 & 72.2 \\
        CORAL(No Markdown) & 54.4 & 57.0 & 60.2 \\

        \hline
        \end{tabularx}
        \begin{tablenotes}
             \item Training on markdown data in addition to code significantly increases performance independent of maximum sequence length.
        \end{tablenotes}
        \label{tab:max_seq_length}
        
    \end{threeparttable}
\end{table}
}

\newcommand{\tableCoverage}{
\begin{table}[]
    
    \caption{The impact of weak supervision on the performance of {\modelname}}
    \centering
    \begin{threeparttable}
        \begin{tabular}{c c}
        \hline
        Heuristic Coverage & Accuracy(\%)\\
        \hline
        100\% & 70.77\\
        75\% & 57.99\\
        50\% & 46.25\\
        25\% & 40.17 \\
        \hline
        \end{tabular}
         \begin{tablenotes}
            \item Higher coverage of weak supervision heuristics significantly increases accuracy.
        \end{tablenotes}
    
        \label{tab:coverage}
    \end{threeparttable}
\end{table}
}

\newcommand{\tableTrainSize}{
\begin{table}[]
    \centering
    \begin{threeparttable}
        \vspace{-10pt}
         \caption{\modelname~accuracy across various training dataset sizes}
        \begin{tabularx}{0.99\columnwidth}{@{\extracolsep{\fill}}l c c c c}
        \hline
         
         Model & \multicolumn{4}{c}{ Number of Cells}  \\
           & 1k & 10k & 100k & 1M  \\
        \hline
        CORAL                       & 61.9 & 62.7 & 63.6 & 72.2 \\
        CORAL (No Masked Attention)  & 53.6 & 57.0 & 59.7 & 70.4 \\
        BERT (AST, No Finetuning)   & 41.0 & 52.4 & 63.2 & 67.9 \\
        
        \hline
        \end{tabularx}
        \begin{tablenotes}
            \item Performance consistently increases with more training data but remains promising even with three orders of magnitude less training data.
        \end{tablenotes}
        \vspace{-10pt}
        \label{tab:train_size}
    \end{threeparttable}
\end{table}
}

\newcommand{\tableWeakSupervision}{
\begin{table}[t]
    \centering
    \begin{threeparttable}
        \vspace{-10pt}
         \caption{\modelname~accuracy across weak supervision coverage}
        \begin{tabularx}{0.99\columnwidth} {@{\extracolsep{\fill}}l c c c} 
           \hline
             Model & \multicolumn{3}{c}{Weak Supervision Coverage} \\
             & 25\% & 50\% & 100\% \\
             
            \hline
            CORAL & 41.6 & 46.4 & 72.2 \\
            CORAL(No Masked Attention) & 31.7 & 46.1 & 70.4 \\
            BERT(AST, No Finetuning) & 26.1 & 47.2  & 67.9\\
            \hline
        \end{tabularx}
        \begin{tablenotes}
            \item Training with more weak supervision significantly improves performances. 
        \end{tablenotes}
        \vspace{-10pt}
        \label{tab:weak_super}
    \end{threeparttable}
\end{table}
}

\newcommand{\tableHoldOutSeeds}{
\begin{table*}[]
    \centering
    \begin{threeparttable}
     \caption{Fraction of predicted stages for cells that contain previously unseen functions}
    \begin{tabular}{l l c c c c c}
    \hline
     Function 											& Expectation		& \IMPORT & \WRANGLE & \EXPLORE & \MODEL & \EVALUATE    \\                
    \hline                      		
    \textit{pandas.DataFrame.dropna} 							& Wrangle					& 0 & \textbf{0.93}&	0.07&	0 &	0        \\          
    \textit{pandas.DataFrame.groupby} 							& Wrangle					& 0 & \textbf{0.52}	& 0.12&	0.02  &	0.34     \\            
    \textit{seaborn.jointplot} 									& Explore					& 0 & 0.00 &	\textbf{0.98}&	0.00&	0.02 \\               
    \textit{seaborn.countplot} 									& Explore					& 0 & 0.01 &	\textbf{0.98}&	0.00&	0.01 \\         
    \textit{sklearn.linear\_model.SGDClassifier} 				& Model					& 0 & 0 &	0&	\textbf{0.67}&	0.32             \\       
    \textit{sklearn.linear\_model.PassiveAggressiveClassifier} 	& Model					& 0 & 0.06&	0&	\textbf{0.61}&	0.39    \\
    \textit{sklearn.metrics.f1\_score} 							& Evaluate					& 0 & 0&	0.01&	0.05& \textbf{0.94}      \\      
    \textit{sklearn.metrics.log\_loss} 							& Evaluate					& 0.02& 0.01&	0.02&	0.26&	\textbf{0.70} \\
    \hline
    \end{tabular}
    \begin{tablenotes}
        \item {\modelname} accurately categorizes common data analysis functions as frequently belonging to their expected stage.
    \end{tablenotes}
    \end{threeparttable}
    \label{tab:hold_out_seeds}
\end{table*}{}
}

\newcommand{\figErrorAnalysis}{
\begin{figure}[t]
    \includegraphics[width=\columnwidth]{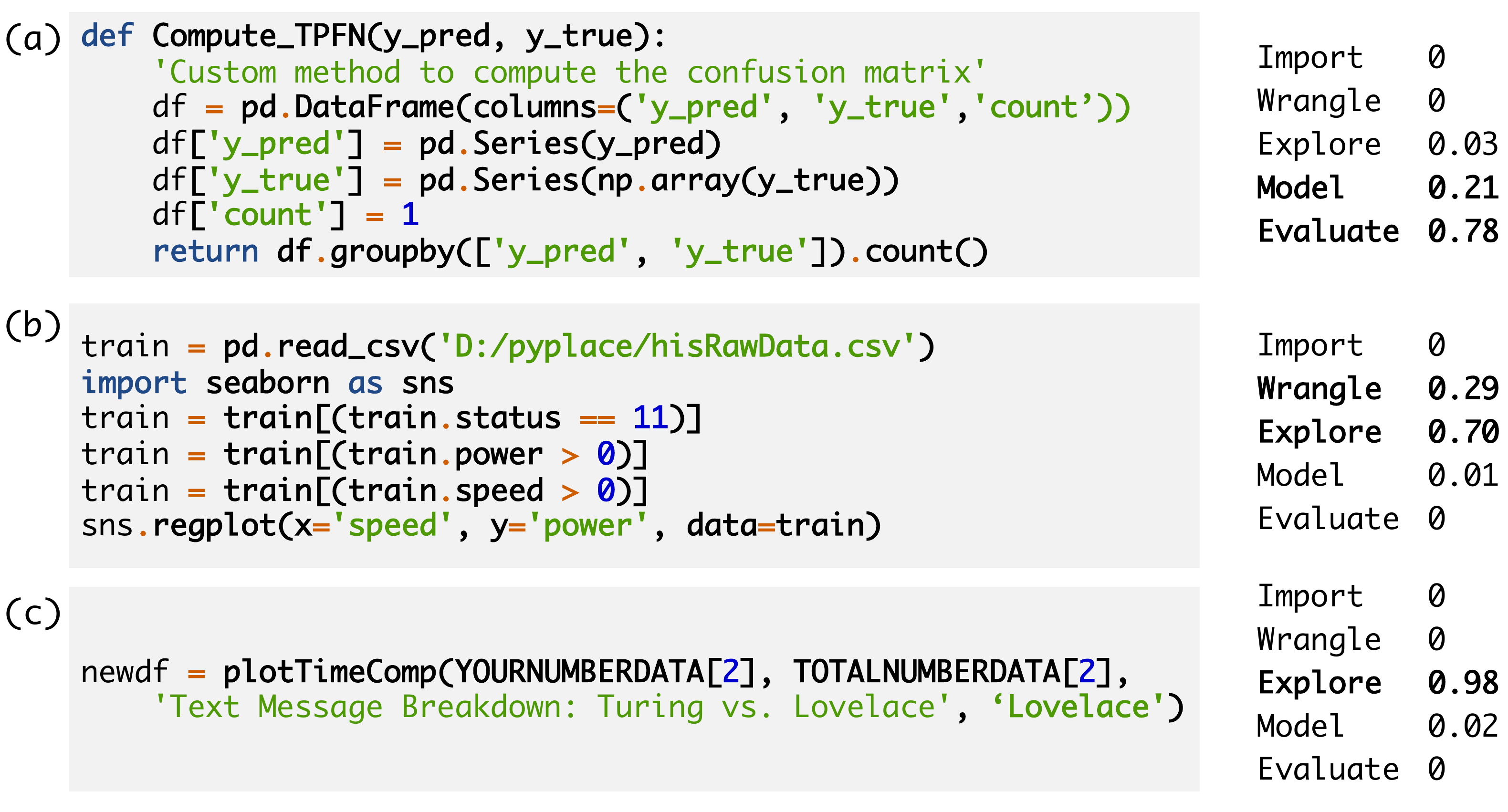}
    \vspace{-22pt}
   \caption{Example predictions. Probability distributions over stages from CORAL's SoftMax output (Eq.~\eqref{eq:topic}) are listed on the right side. In (a), CORAL correctly identifies the cell as \EVALUATE\ rather than \WRANGLE , likely by interpreting "confusion matrix", perhaps based on previously seen markdown. In (b), the model identifies the use of \texttt{sns.regplot}, an unseen statistical visualization function, as an example of \EXPLORE. In (c), CORAL correctly interprets a user-defined function.}
    \vspace{-15pt}
    \label{fig:error_analysis}
\end{figure}
}

\newcommand{\figCombined}{
\begin{figure}
    \centering
    \begin{minipage}[b]{0.45\textwidth}
        \centering
        \includegraphics[width=\textwidth]{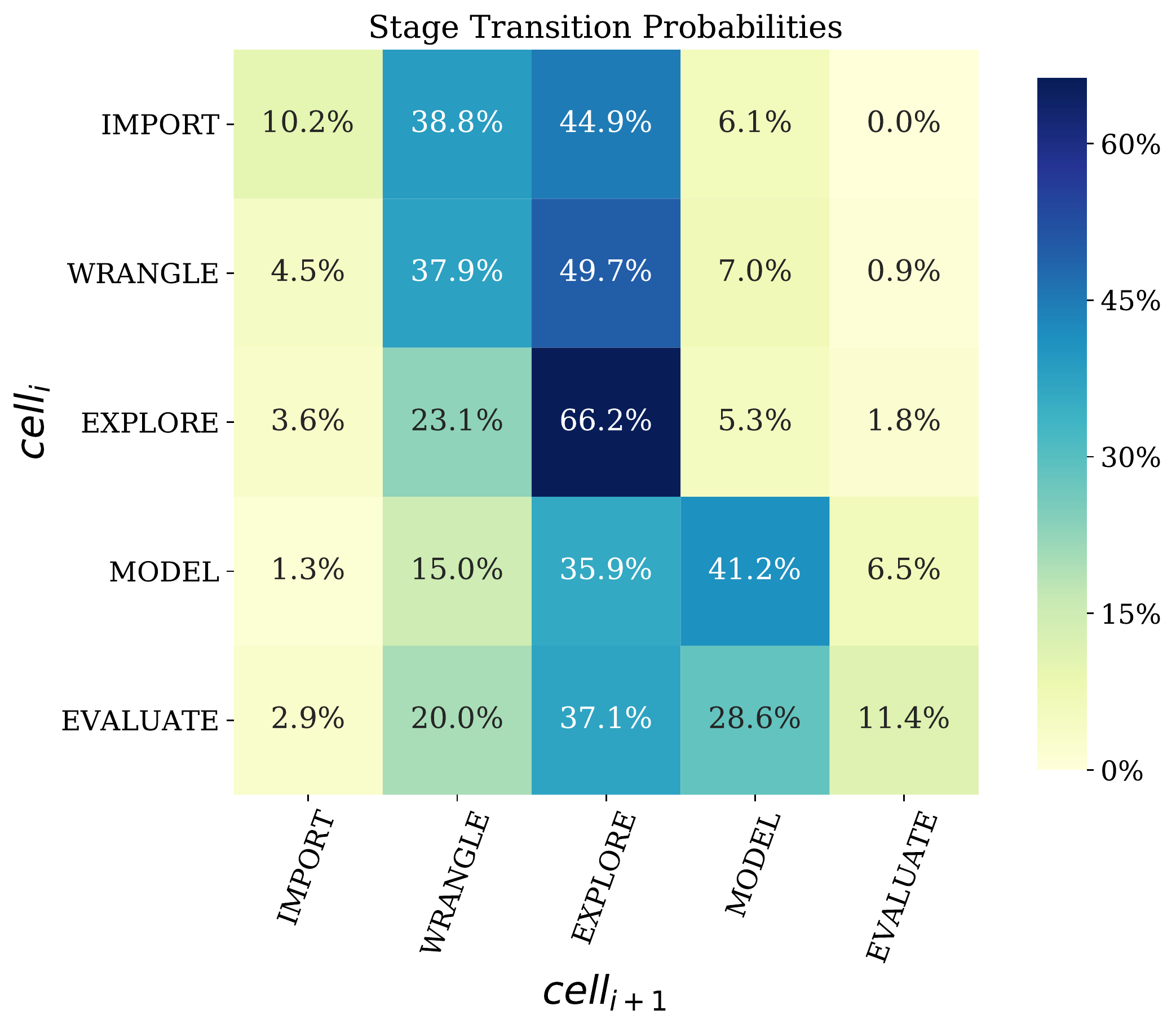}
        \caption{Transitions between data science stages in a corpus of 118k Jupyter Notebooks.}
        \label{fig:transitions}
    \end{minipage}\hfill
    \begin{minipage}[b]{0.45\textwidth}
        \centering
        \includegraphics[width=0.9\textwidth]{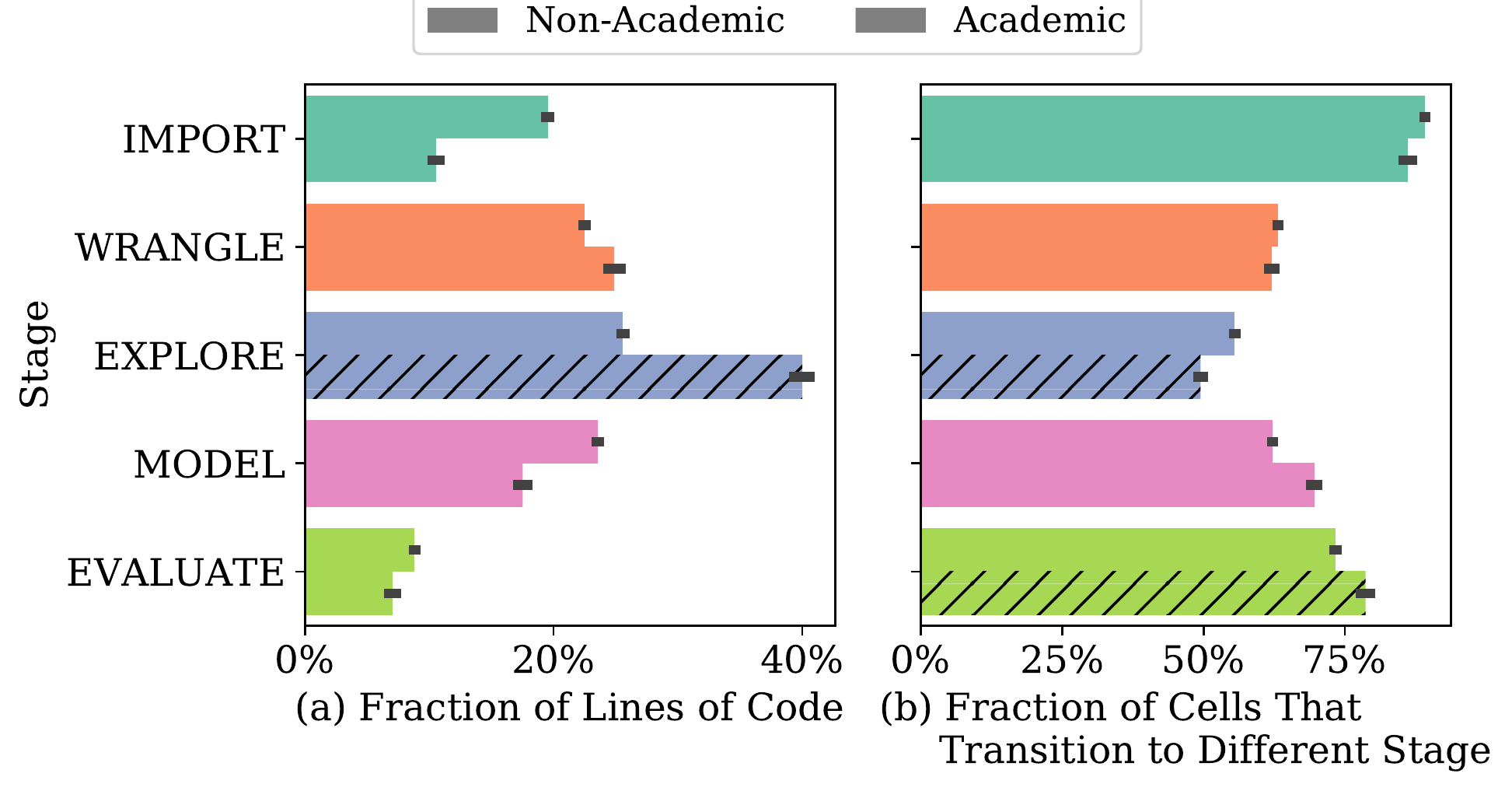}
        \caption{The fraction of code devoted to each stage.}
        \label{fig:frac_code_per_stage}
        \end{minipage}
\end{figure}
}

\newcommand{\figTransition}{
\begin{figure}[t]
    \vspace{-12pt}
    \centering
    \includegraphics[width=0.35\textwidth]{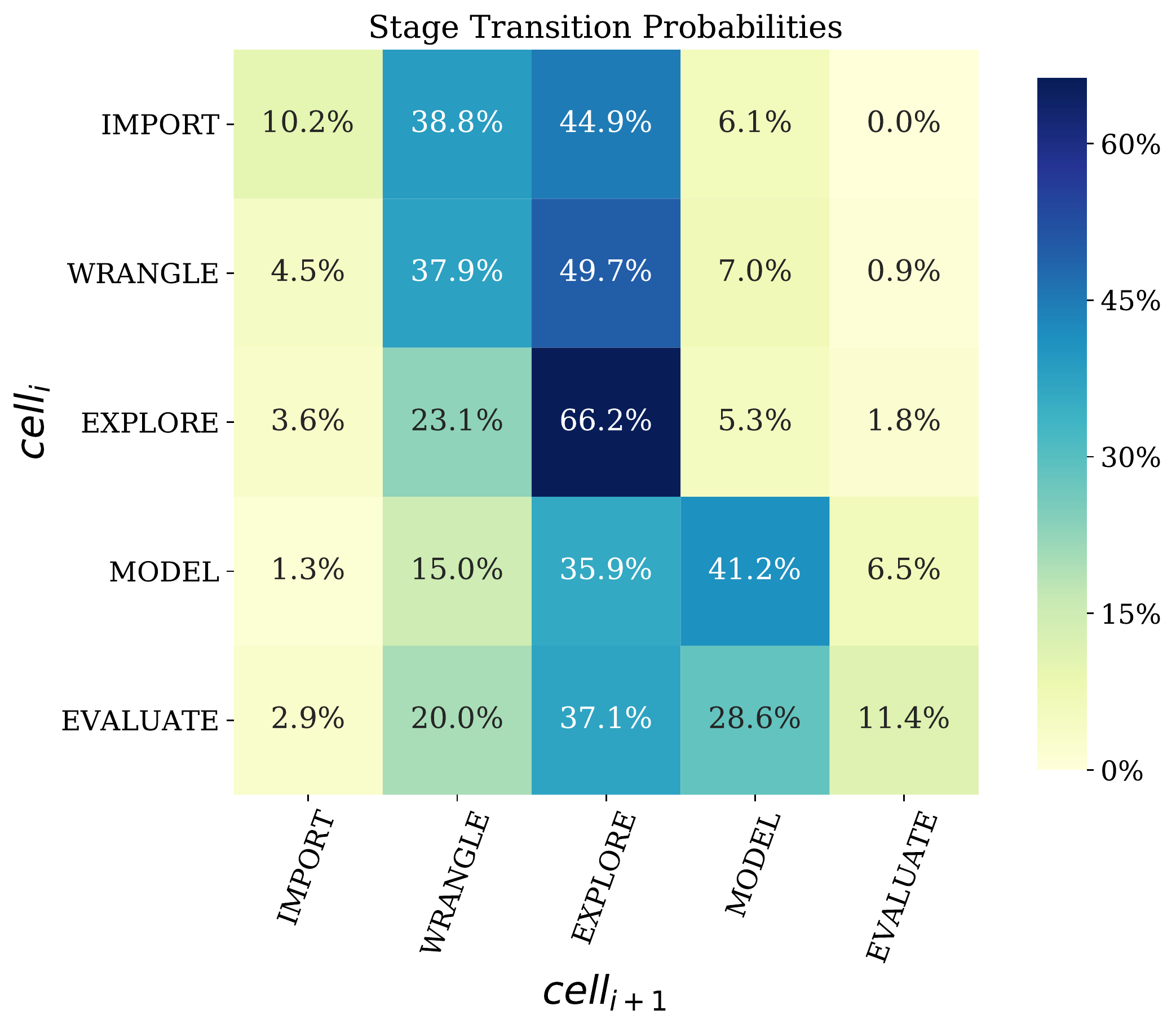}
    \vspace{-12pt}
    \caption{Transitions between data science stages in a corpus of 118k Jupyter Notebooks.}
    \label{fig:transitions}
\end{figure}
}

\newcommand{\figFracCode}{
\begin{figure}[t]
    \vspace{-15pt}
    \centering
    \includegraphics[width=0.99\columnwidth]{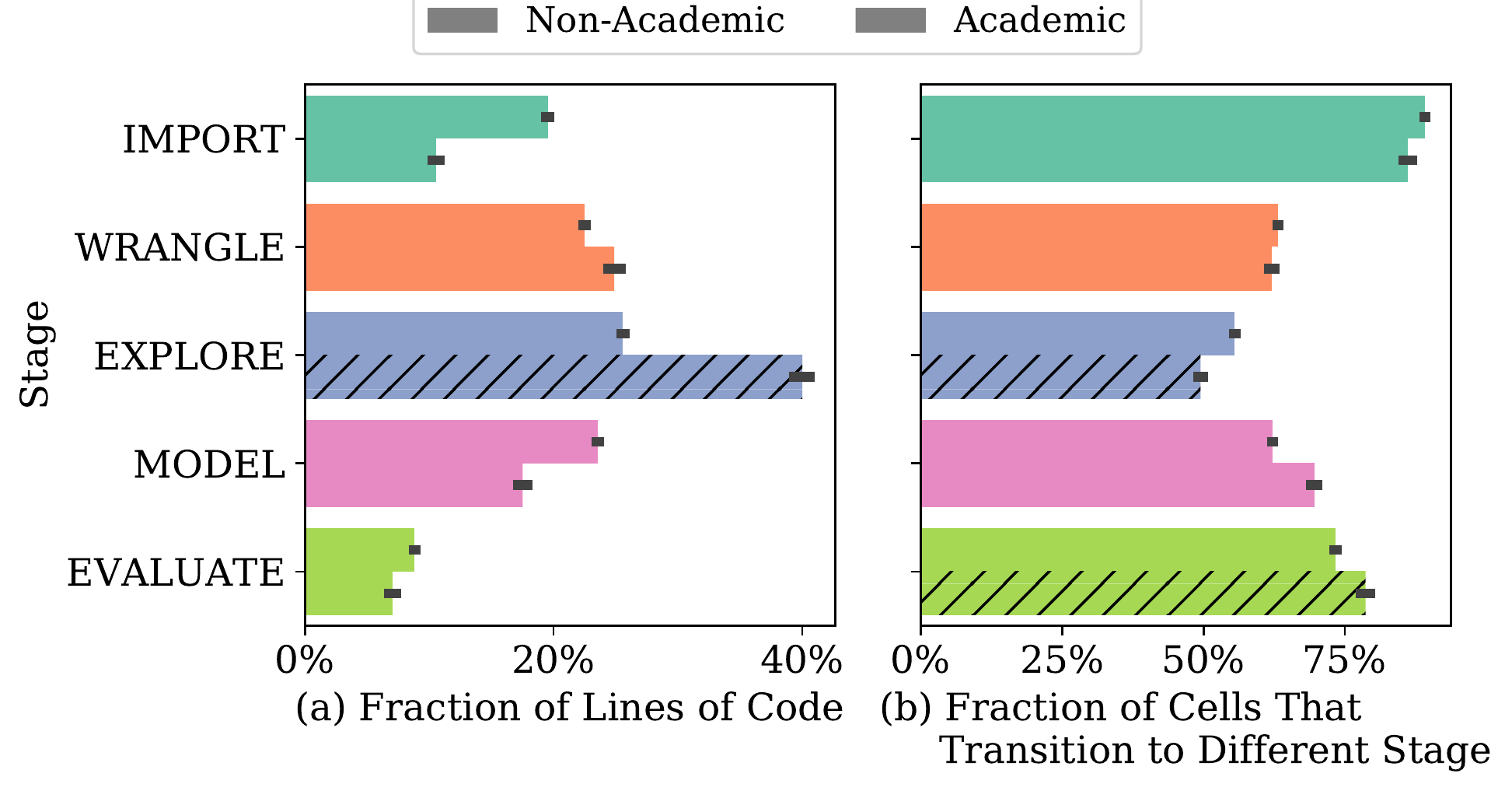}
    \vspace{-20pt}
    \caption{Differences between academic and non-academic notebooks.}
    \label{fig:frac_code_per_stage}
    \vspace{-15pt}
\end{figure}
}

\newcommand{\figPerDomain}{
\begin{figure}
    \centering
    \vspace{-15pt}
    \includegraphics[width=0.9\columnwidth]{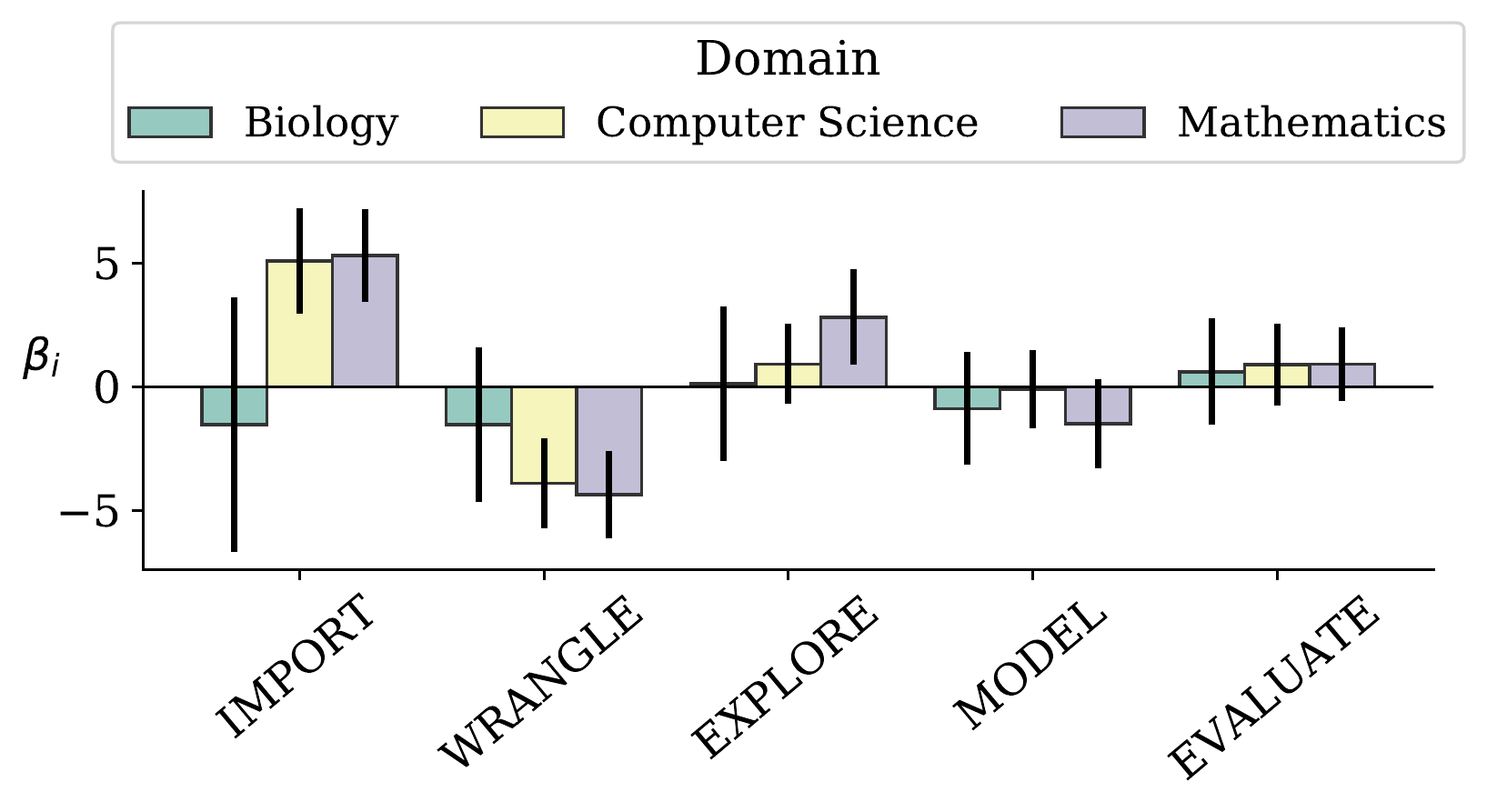}
    \vspace{-10pt}
    \caption{Results from (R2), indicating differences in how paper impact in different domains is related to the content of associated notebooks.}
    \vspace{-15pt}
    \label{fig:per_domain_betas}
\end{figure}
}

\newcommand{\figConfusion}{
\begin{figure}[ht]
    \centering
    \includegraphics[width=0.4\columnwidth]{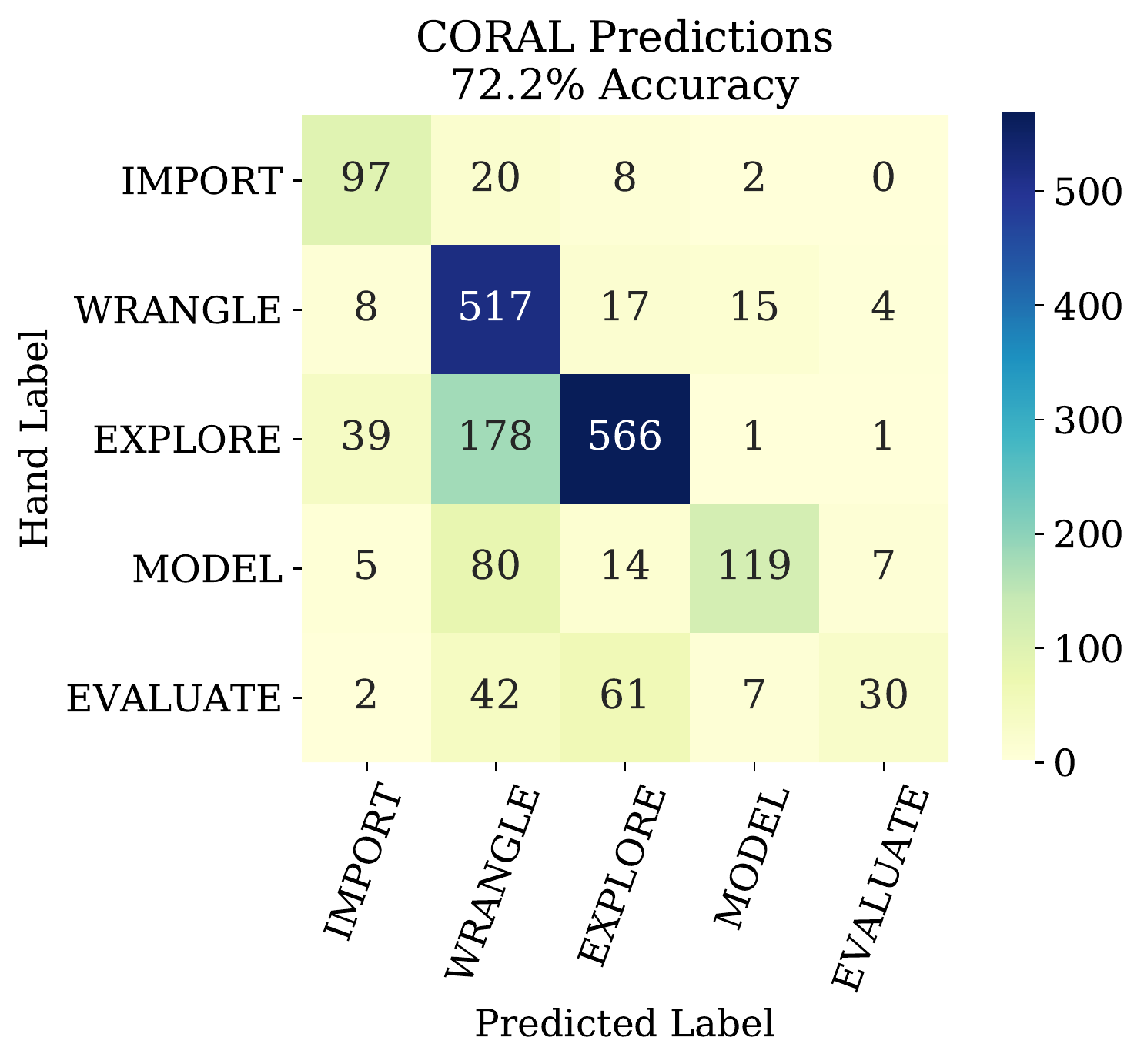}
    \vspace{-15pt}
    \caption{Confusion matrix for \modelname's predictions on the data analysis stage prediction task.}
    \label{fig:model_confusion}
\end{figure}
}

\newcommand{\tableSeed}{
\begin{center}
\begin{table}[ht]
   \caption{Seed functions with associated data analysis stages used in weak supervision heuristics (Section~\ref{sec:weak_supervision}).}
  \centering
  \begin{tabular}{p{0.25\columnwidth}p{0.5\columnwidth}} 
    \hline
    Stage & Seed Functions \\
    \hline
    \multirow{9}{*}{Wrangle} & pandas.read\_csv \\
     & pandas.read\_csv.dropna \\
     & pandas.read\_csv.fillna \\
     & pandas.DataFrame.fillna \\
     & sklearn.datasets.load\_iris \\
     & scipy.misc.imread \\
     & scipy.io.loadmat \\
     & sklearn.preprocessing.LabelEncoder \\
     & scipy.interpolate.interp1d \\
     
     \hline
     
     \multirow{8}{*}{Explore} & matplotlib.pyplot.show\\
     & matplotlib.pyplot.plot\\
     & matplotlib.pyplot.figure\\
     & seaborn.pairplot\\
     & seaborn.heatmap\\ 
     & seaborn.lmplot\\
     & pandas.read\_csv.describe\\
     & pandas.DataFrame.describe \\
    
    \hline 
    
    \multirow{14}{*}{Model} & sklearn.decomposition.PCA\\
     & sklearn.naive\_bayes.GaussianNB\\               
     & sklearn.ensemble.RandomForestClassifier\\
     & sklearn.linear\_model.LinearRegression\\
     & sklearn.linear\_model.LogisticRegression\\
     & sklearn.tree.DecisionTreeRegressor\\
     & sklearn.ensemble.BaggingRegressor\\
     & sklearn.neighbors.KNeighborsClassifier\\
     & sklearn.naive\_bayes.MultinomialNB\\
     & sklearn.svm.SVC\\
     & sklearn.tree.DecisionTreeClassifier\\
     & tensorflow.Session\\
     & sklearn.linear\_model.Ridge\\
     & sklearn.linear\_model.Lasso \\
     
     \hline
     
    \multirow{5}{*}{Evaluate} &  sklearn.cross\_validation.cross\_val\_score\\
     & sklearn.metrics.mean\_squared\_error\\
     & sklearn.model\_selection.cross\_val\_score\\ 
     & scipy.stats.ttest\_ind\\ 
     & sklearn.metrics.accuracy\_score\\
     
     \hline

    \end{tabular}
    \label{tab:seed-functions}
    \vspace{-20pt}
\end{table}
\end{center}

}

\newcommand{\tableRubric}{
\begin{center}

\begin{table*}[ht]
    \caption{Qualitative rubric used for labeling the Expert Annotated Dataset (Section~\ref{sec:expert_annotated_notebooks}) used for final model evaluation.}
  \centering
  \begin{tabular*}{\textwidth}{p{0.05\textwidth}p{0.2\textwidth}p{0.175\textwidth}p{0.175\textwidth}p{0.3\textwidth}} 
    \hline
    Stage & Definition & When to Use & When Not to Use& Example\\
    
    \hline
    Import & These cells are used primarily to import libraries into the Python environment. Although they may serve other functions, like defining constants or initializing helper objects, the majority of the code in these cells sets up analytical tools for use later in the notebook. &  Loading libraries, defining constants, initializing environments, connecting to databases & A cell has one or more import statements, but most of the cell serves another purpose
    & \raisebox{-\totalheight}{\includegraphics[width=0.3\textwidth]{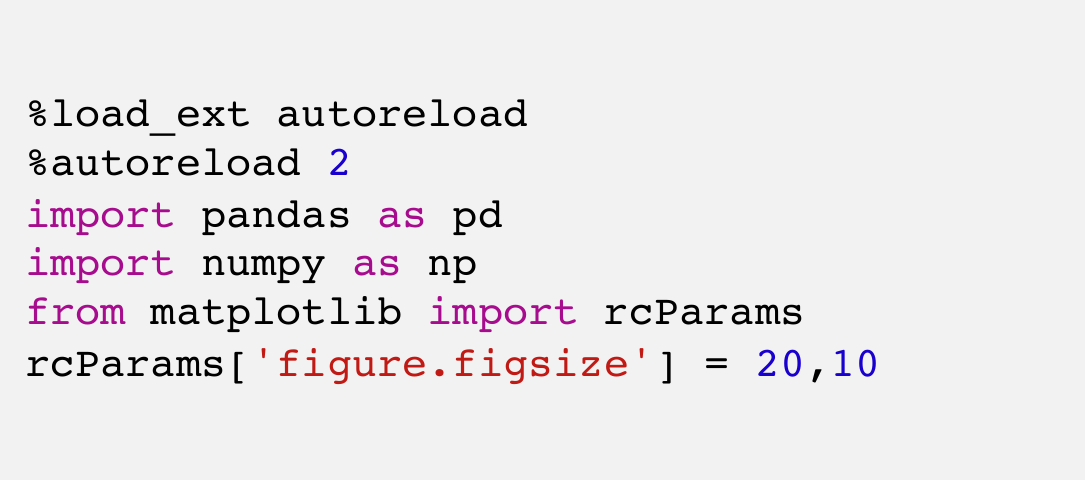}}\\
    
    \hline
     Wrangle & Wrangle cells clean, filter, summarize, and/or integrate data. These cells often permute data for use in later cells. &  Cleaning data, feature processing, data transformations, augmenting an existing dataset, loading and/or saving data, splitting data into train and test sets & Transformations are applied, but the result is simply examined (See: Explore)
    & \raisebox{-\totalheight}{\includegraphics[width=0.3\textwidth]{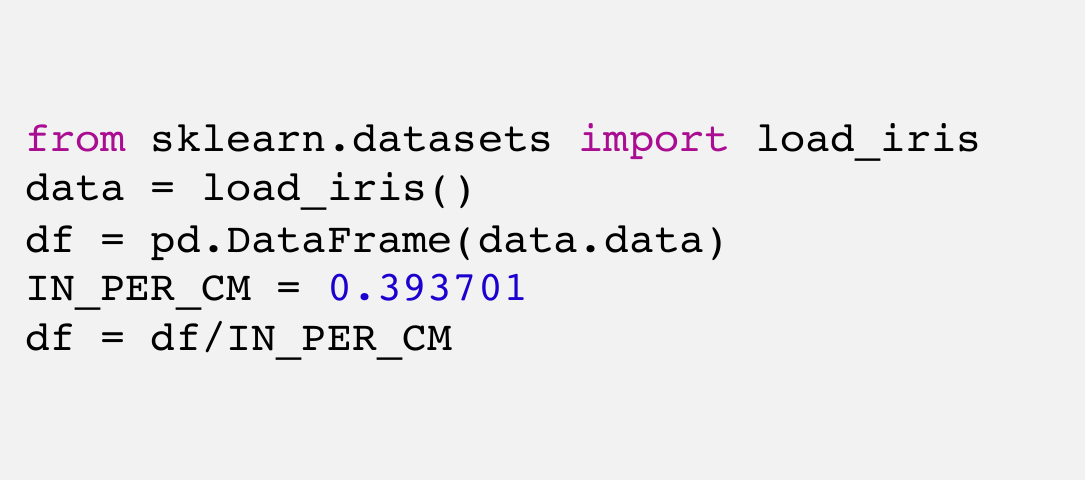}}\\
    
    \hline
    Explore & Interactive explorations of data. These cells tend to yield a result that informs later decisions, or enable the user to draw new conclusions. Explore cells may also transform data, but only for the purpose of exploring relationships and not for further in-depth analysis &  Rendering DataFrames, visualizing relationships, printing summaries of data, calculating simple statistics, examining the output of functions & Visualizations are used to evaluate the performance of a model (See: Evaluate)
    & \raisebox{-\totalheight}{\includegraphics[width=0.3\textwidth]{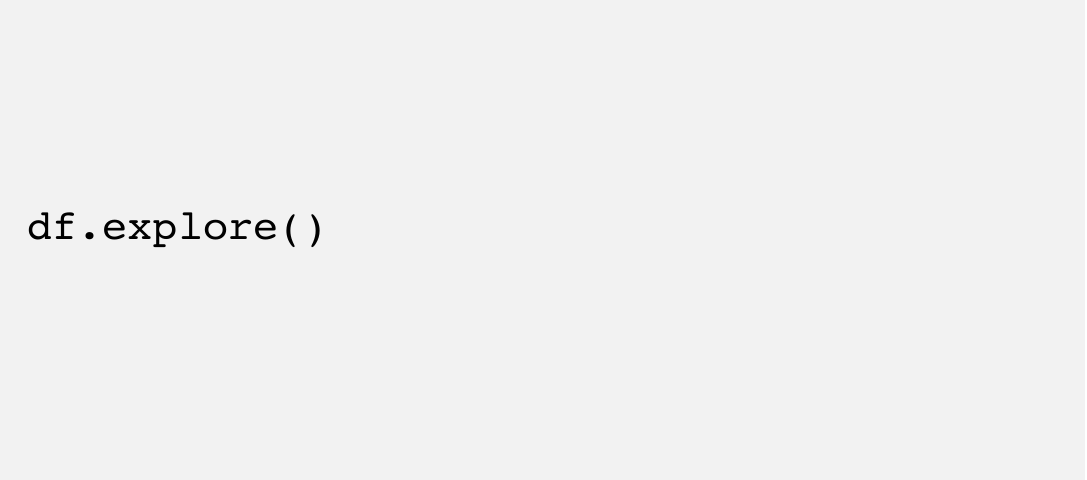}}\\
    
    \hline
    Model & Define and fit models of relationships to data. These cells may include some data transformations, but the primary purpose is to create a model to describe or predict some facet of the dataset &  Statistical modeling, fitting and/or specifying machine learning models, simulation, defining loss functions & Significance testing and calculating feature importance (See: Evaluate)
    & \raisebox{-\totalheight}{\includegraphics[width=0.3\textwidth]{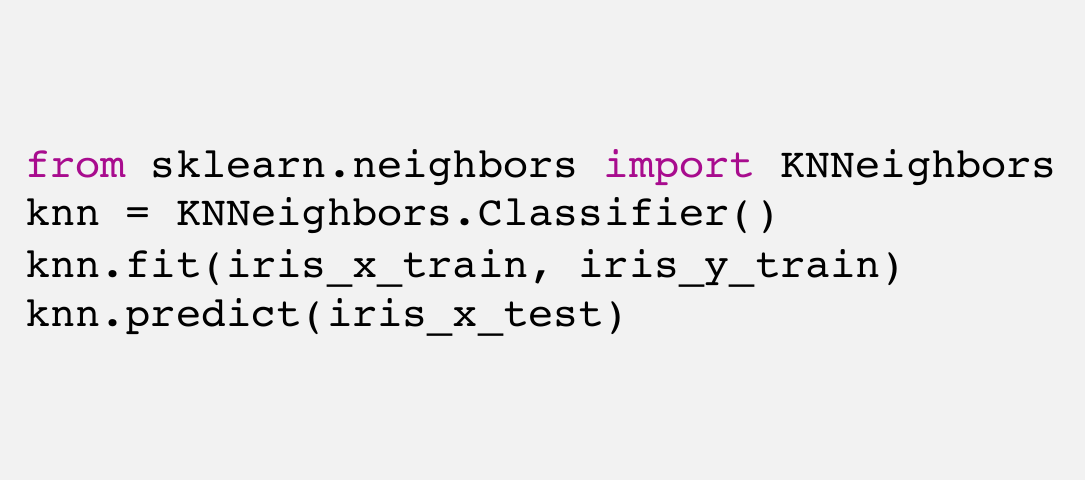}}\\
    
     \hline
    Evaluate & Measure the explanatory power or predictive accuracy of model using appropriate statistical techniques. These cells sometimes employ visualizations to explore analytical results (e.g. plotting regression residuals) & Cross validation, significance testing, inspecting model output, plotting feature significance. & If a cell both evaluates and defines a machine learning model (a common pattern), default to "Model"
    & \raisebox{-\totalheight}{\includegraphics[width=0.3\textwidth]{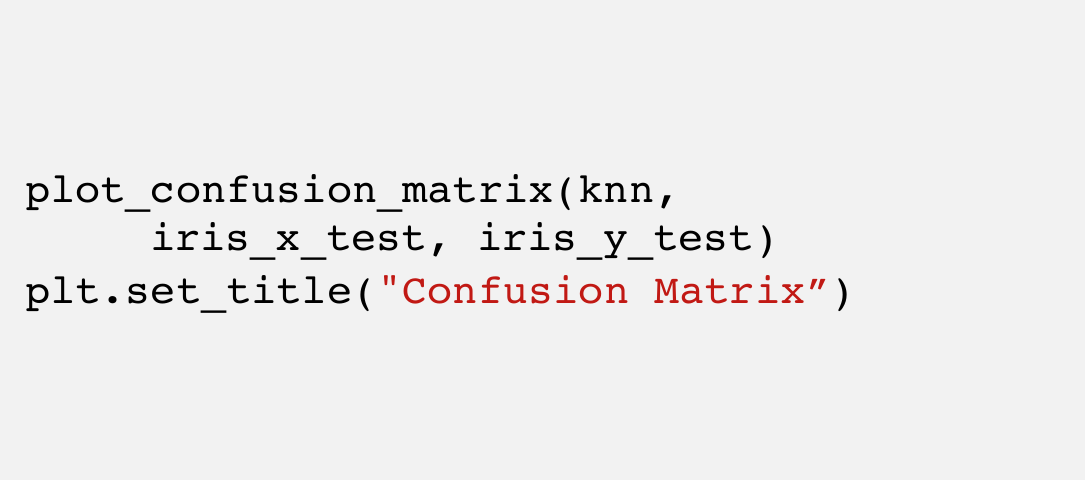}}\\
    
    \hline
    \end{tabular*}
    
    \label{tab:rubric}
\end{table*}
\end{center}
}
\section{Introduction}
\label{sec:intro}


Data analysis is central to the scientific process. Increasingly, analytical results are derived from code, often in the form of computational notebooks, such as Jupyter notebooks~\cite{kluyver2016jupyter}. 
Analytical code is becoming more frequently published in order to improve replication and transparency \cite{foster_open_2017,ayers_libguides_nodate,pradal_publishing_2013}.
However, as of yet no tools exist to study unlabeled source code both at scale and in depth. 
Previous in-depth analyses of scientific code heavily rely on expert annotations, limiting the scale of these studies to the order of a hundred examples \cite{liu_paths_nodate,rule_exploration_2018}.
Large-scale studies across thousands of examples have been limited to simple summaries such as the number or nature of imported libraries, total line counts, or the fraction of lines that are used for comments \cite{rule_exploration_2018,rehman_towards_2019,wang_better_2019}. The software engineering community has emphasized the inadequacy of these analyses, noting that \say{there is a strong need to programmatically analyze Jupyter notebooks}~\cite{wang2019_better}, while HCI researchers have observed that studying the data science process through notebooks may play a role in addressing the scientific reproducability crisis \cite{liu_paths_nodate,kery_story_2018}.



Automated annotation tools could enable researchers to answer important questions about the scientific process across millions of code artifacts. Do analysts share common sequential patterns or processes in their code? Do different scientific domains have different standards or best practices for data analysis? How does the content of scientific code relate to the impact of corresponding publications?
To draw insights on the data science process, previous work has conceptualized the analysis pipeline as a sequence of discrete stages starting from importing libraries and wrangling data to evaluation~\cite{kandel_enterprise_2012, wongsuphasawat2019, alspaugh2018}.
Building on this conceptual model, our goal is to develop a tool that can automatically annotate code blocks with the analysis stage they support, enabling large-scale studies of scientific data analysis to answer the questions above.

Analyzing scientific code is particularly difficult because as a \say{means to an end}~\cite{johanson_software_2018}, scientific code is often messy and poorly documented.
Researchers engage in an iterative process as they transition between tasks and update their code to reflect new insights \cite{kery_variolite_2017,hill_trials_2016}.
As such, a computational notebook may interleave snippets for importing libraries, wrangling data, exploring patterns, building statistical models, and evaluating analytical results, thereby building a complex and frequently non-linear sequence of tasks \cite{liu_paths_nodate,kandel_enterprise_2012}.
While some analysts use markdown annotations, README's, or code comments to express the intended purpose of their code, these pieces of documentation are often sparse and rarely document the full analysis pipeline \cite{rule_exploration_2018}. Domain-specific best practices, techniques, and libraries may additionally obfuscate the intent of any particular code snippet.
As a result, interpreting scientific code typically requires significant expertise and effort, making it prohibitively expensive to obtain ground truth labels on a large corpus, and therefore infeasible to build annotation tools which require anything more than minimal supervision.


In this paper, we present {\modelnamelong} ({\modelname}) to classify scientific code into stages in the data analysis process. 
Importantly, the model requires only easily available weak supervision in the form of five simple heuristics, and does not rely on any manual annotations. We show that CORAL learns new relationships beyond the information provided by these heuristics, indicating that currently popular transformer architectures~\cite{vaswani2017attention} can be extended to weakly supervised tasks with the addition of a small amount of expert guidance. Our model achieves high agreement with human expert annotators and can be scaled to analyze millions of code artifacts, uniquely enabling large-scale studies of scientific data analysis.

We describe a new task for classifying code snippets as stages in the data analysis process (\sect{sec:task_description}). We provide an extension to a corpus of 1.23M Jupyter Notebooks (\sect{sec:jupyer_notebooks}): a new dataset of expert annotations of stages in the data analysis process for 1,840 code cells in 100 notebooks, which we use exclusively for evaluation and \emph{not} for training (\sect{sec:expert_annotated_notebooks}).

Next we describe {\modelname} (\sect{sec:methods}): a novel graph neural network model for embedding data science code snippets and classifying them as stages in the data science process. To capture semantic clues about the analyst's intention, \modelname\ uses a novel masked attention mechanism to jointly model natural language context (such as markdown comments) with structured source code (\sect{subsec:encoder_attention}). We implement a weakly supervised architecture with five simple heuristics to compensate for the absence of labels, as labeling code requires domain expertise and is therefore expensive and infeasible at massive scale (\sect{sec:weak_supervision}). To further compensate for limited labels, \modelname\ combines this weak supervision with unsupervised topic modeling into a multi-task optimization objective (\sect{subsec:optimization_objective}).

We evaluate our model (\sect{sec:evaluation}) by comparing it to baselines including expert heuristics, weakly supervised LDA, and state-of-the-art neural representation techniques (\sect{sec:baselines}). We demonstrate that {\modelname}, using both code and surrounding natural language annotations, outperforms expert heuristics by 36\% and significantly outperforms all other baselines. Through an ablation study we demonstrate that increased maximum sequence length $M$, weak supervision and unsupervied topic modeling all strictly improve performance, and that including markdown improves performance on cells \textit{without associated markdown} by 13\% (\sect{sec:ablation}).
Further, we explore the impact of maximum input size and dataset size on our model's performance (\sect{sec:additional_experiments}), showing that \modelname\ significantly outperforms all baselines even when trained on only 1k  examples.
In a comprehensive error analysis, we demonstrate that previously unseen data science functions are correctly labeled with appropriate analysis stages (\sect{sec:error_analysis}). 
 
We then deploy our model to resolve previously unanswered questions about data analysis by linking academic notebooks and associated publications to conduct the largest ever study of scientific code (\sect{sec:large_scale_studies}). We find that (1) there are significant differences between academic and non-academic papers, (2) that papers which include references to notebooks receive on average 22 times the number of citations as papers that do not, and (3) that papers linked to notebooks that more evenly capture the full data science process in expectation receive twice the number of citations for every one standard deviation increase in entropy between stages. 

In summary, the contributions of this paper are:
\begin{itemize}
    \item A new task and public dataset for classifying Jupyter cells as stages in the data science process (\sect{sec:datasets}).
    \item A multi-task, weakly supervised transformer architecture for classifying code snippets which jointly models natural language and code (\sect{sec:methods}).
    \item A comprehensive evaluation of code representation learning methods (\sect{sec:evaluation}).
    \item The largest ever study of scientific code (\sect{sec:large_scale_studies}).
\end{itemize}
We make all code and data used in this work publicly available at {\projectsite}. 
\label{sec:related_work}
\section{Related Work}

\subsection{{Representation Learning for Source Code}}
\label{subsec:related_work_rep_learning}
Early methods for code representational learning treated source code as sequence of tokens and built language models on top~\cite{hindle2012naturalness, tu2014localness, nguyen2013statistical, allamanis2013mining, raychev2014code}.
Later work incorporated additional information specific to source code, such as object-access patterns~\cite{kwon_software_behavior}, code comments~\cite{movshovitz2013natural}, parse trees~\cite{bielik2016phog}, serialized Abstract Syntax Trees (ASTs)~\cite{alon2018general, li2017code},  ASTs as graph structures~\cite{allamanis2017learning}, and associated repository  metadata~\cite{HiGitClass9}.
As documentation (in markdown format) is prevalent in Jupyter notebooks~\cite{rule_exploration_2018}, our model incorporates both markdown text and graph-structured ASTs, taking advantage of both semantic and structural information.

Due to the scarcity of labeled examples, most previous work learned code representations without supervision~\cite{alon2019code2vec, allamanis2014learning, acharya2007mining, nguyen2017exploring, raychev2014code}.
The learned representations were mostly used for hole completion tasks, including the prediction of self-defined function names~\cite{alon2019code2vec}, API calls~\cite{acharya2007mining, nguyen2017exploring}, and variable names~\cite{allamanis2014learning, raychev2014code}.
In contrast, our task -- classifying code cells as analysis stages -- arguably requires a higher level understanding of the intention of code. 
To overcome the bottleneck of manual labeling, we turn to weak supervision.
\textit{Snorkel}~\cite{ratner2019snorkel} combined labels from multiple weak supervision sources, denoised them, and used the resulting probabilistic labels to train discriminative models.
Building on this idea, we introduce weak supervision to code representation learning by leveraging a small number of expert-supplied heuristics. 

\subsection{Graph Neural Networks}
\label{subsec:related_work_GNN}
GNNs are powerful tools for a variety of tasks, including node classification~\cite{hamilton2017inductive, kipf2016semi}, text classification~\cite{wuGNNTextClassification}, link prediction~\cite{schlichtkrull2018modeling, zhang2018link}, graph clustering~\cite{defferrard2016convolutional, ying2018hierarchical} and graph classification~\cite{ying2018hierarchical, dai2016discriminative, duvenaud2015convolutional, scarselli2008graph}. Additional work suggests that feeding  underlying graphical syntax to a natural language model can improve generalization~\cite{battaglia2018relational}. Tree structures have been show to help summarize source code \cite{fernandes2018structured}, and complete code snippets ~\cite{allamanis2017learning, Brockschmidt2018GenerativeCM} in code representation learning. We build on prior work in attention-based graph neural networks~\cite{velivckovic2017graph} and adopt a self-attention mechanism in our model that jointly learns from ASTs and markdown text.

\subsection{Studies of Data Analysis Practices}
\label{subsec:related_work_data_analysis_practtices}
There is significant existing research on understanding data analysis practices (\eg \cite{kandel_enterprise_2012, alspaugh2018, kery_story_2018, kery_variolite_2017, liu_paths_nodate}), mostly using qualitative methods to elicit experiences from analysts.
Some interviews focused specifically on Jupyter notebook users~\cite{kery_story_2018, rule_exploration_2018}.
Despite synthesizing rich observations, interview studies were limited to dozens of participants.
A few studies conducted large-scale analysis of Jupyter notebooks, but were limited to simple summary statistics~\cite{rule_exploration_2018}, a single library~\cite{rehman_towards_2019}, or code quality~\cite{wang_better_2019}.
Our model enables the analysis of data science both at scale and in depth, which may validate and complement findings from previous qualitative studies. 

\subsection{The Data Science Process}
\label{subsec:related_work_the_data_science_process}
A related branch of work~\cite{kandel_enterprise_2012, alspaugh2018, wongsuphasawat2019} modeled the data analysis process as a sequence of iteratively visited stages. Other authors have noted that a better understanding of this process could improve scientific reproducability \cite{liu_paths_nodate}, aid in the development of new analysis tools~\cite{kery_variolite_2017,kery_story_2018}, and identify common points of failure \cite{li_asr_2013}.

\section{Prediction Task \& Datasets}
\label{sec:datasets}
We present a new task for labeling code snippets as stages in the data science process (Figure~\ref{fig:task_description}), identify a corpus of computational notebooks for large-scale training, and provide a new dataset of expert annotations that are used exclusively in the final evaluation. 

\figTask
\subsection{Prediction Task}
\label{sec:task_description}

In order to automatically learn useful data science constructs from code, we propose a new task and accompanying dataset for classifying code snippets as stages in the data science process. Figure~\ref{fig:task_description} shows five mock examples from this task. We task models with associating a snippet with one of five labels, which are drawn from and motivated by previous work: \IMPORT, \WRANGLE, \EXPLORE, \MODEL, and \EVALUATE\ (\sect {subsec:related_work_the_data_science_process}). \IMPORT\ cells primarily load external libraries and set environment variables, while \WRANGLE\ cells load data and perform simple transformations. \EXPLORE\ cells are used to visualize data, or calculate simple statistics. \MODEL\ cells define and fit statistical models to the data, and finally \EVALUATE\ cells measure the explanatory power and/or significance of models. Additional details on these stages is available in Appendix Table~\ref{tab:rubric}.

\subsection{Jupyter Notebook Corpus for Training}
\label{sec:jupyer_notebooks}
We curate a training set for this task by building upon the UCSD Jupyter notebook corpus, which contains all 1.23M publicly available Jupyter notebooks on Github \cite{rule_exploration_2018}. 
Jupyter is the most popular IDE among data scientists, with more than 8M users \cite{jetbrainsDS,kelley2017jupyter}, at least in part because it enables users to combine code with informative natural language markdown documentation.  As noted by the corpus' authors, the dataset contains many examples of the myriad uses for notebooks, including completing homework assignments, demonstrating concepts, training lab members, and more~\cite{rule_exploration_2018}. For the purposes of this paper we filtered the corpus to those notebooks that transform, model, or otherwise manipulate data by limiting our analysis to notebooks that import  \textit{pandas}, \textit{statsmodels}, \textit{gensim}, \textit{keras}, \textit{scikit-klearn}, \textit{xgboost} or \textit{scipy}. This leaves us with a total of 118k Jupyter notebooks, which we randomly split into training (90\%) and validation sets (10\%). 
These notebooks are not annotated with any ground truth labels of data science stages.
Thus, we propose a combination of unsupervised representation learning and weak supervision to study them at scale (\sect{sec:methods}).
    
\subsection{Expert Annotated Notebooks (Only Used for Evaluation)}
\label{sec:expert_annotated_notebooks}
\xhdr{Annotation} We randomly sampled 100 notebooks containing 1840 individual cells from the filtered dataset for hand-labeling. The first two authors, who have significant familiarity with the Python data science ecosystem, independently annotated the cells with one of the five data science stages. The annotators performed a preliminary round of coding, discussed their results, and produced a standardized rubric for qualitative coding, which is available in the \ Appendix \ref{app:qualitative_rubric} (Table \ref{tab:rubric}). The rubric clearly defines each data analysis stage and provides guidelines for when a label should and should not be used. Using this rubric, the annotators each made a second independent coding pass. We evaluate inter-rater reliability with Cohen's kappa statistic, which corrects for  agreement by chance, and find the highest level of correspondence (\say{substantial agreement},  $\kappa=0.803$)~\cite{landis197}.  Finally, the annotators resolved the remaining differences in their labels by discussing each disagreement, producing a final dataset of 1840 cells
for model evaluation (\sect{sec:evaluation}). Our annotation rubric along with all data and code are available at {\projectsite}.
Importantly, these expert annotations are never used in training or validation including model selection, but only for the final evaluation (\sect{sec:evaluation}).

\xhdr{Multi-Class v.s. Multi-Label} Both annotators paid close attention to potentially ambiguous cells while labeling, observing that it was quite rare for a single cell to be used for multiple stages of the data science process (less than 5\% of the time). Furthermore, the median cell in the dataset had two lines of code, making it difficult for a cell to sufficiently express more than one stage. Low label ambiguity at the cell level and high inter-rater reliability support the formulation of this task as multi-class (i.e., five mutually exclusive labels) rather than multi-label (i.e., a cell may have one or more labels), and the selection of cells as the unit of analysis.

 \section{The CORAL Model}
\label{sec:methods}

\figCoral


\modelnamelong{} (\modelname{}) is a model for learning neural representations of data science code snippets and classifying them as stages in the data analysis process. 
{\modelname{}} leverages both source code abstract syntax trees (ASTs) and associated natural language annotations in markdown text  (see \fig{fig:coral}).

\xhdr{Model Contributions} \modelname\ contributes the following:
\begin{itemize}
    \item \modelname\ jointly learns from code and surrounding natural language (\sect{subsec:input_graphs}), while preserving meaningful code structure through a graph-based masked attention mechanism (\sect{subsec:encoder_attention}). We show that adding natural language improves performance by 13\% on snippets that do not have associated markdown comments (\sect{sec:ablation}).
	\item We address the lack of high-quality training data through an easily extensible weakly supervised objective based on five simple heuristics (\sect{sec:weak_supervision}). 
	\item \modelname\ combines this weak supervision with an additional unsupervised training objective (again to avoid costly ground truth labels) based on topic modeling, which we combine with other objectives in a multi-task learning framework (\sect{subsec:optimization_objective}).
\end{itemize} 


\subsection{Input Representations}
\label{subsec:input_graphs}
\modelname\ builds on graph neural networks \cite{gilmer2017neural} and masked-attention approaches \cite{velivckovic2017graph} to encode the AST's graph structure by first serializing the tree and then using its adjacency matrix as an attention mask (\sect{subsec:encoder_attention}). 

We add additional nodes to the AST to capture surrounding natural language. For each code cell, we concatenate its most recent prior markdown as a token sequence to the AST graph sequence (yellow in Figure~\ref{fig:coral}), so long as the markdown is no more than three cells away.
Concretely, we create a node for each markdown token and then connect each markdown node with each AST node.
Finally, we add a virtual node [CLS] (for \textit{classification}) at the head of every input sequence and connect all the other nodes to it. Similar to BERT, we take this node's embedding as the representation of the cell \cite{devlin2018bert}. 


\xhdr{Notation}
Formally, let $\mathcal{V}=\{u,v,...\}$ be the set of nodes in the input, where each node $v$ is either an AST node or markdown token. For any input sequence that has more than $M$ nodes, we truncate it and keep only the first $M$ nodes (a modeling choice which we evaluate in \sect{sec:additional_experiments}). We use $A$ to represent the graph adjacency matrix that encodes the relationship between nodes as described above. 
All input nodes are converted to embedding vectors of dimension $d_{model}$. We assemble these embeddings into a matrix $X$.

\subsection{Encoding Code Cells with Attention}
\label{subsec:encoder_attention}
We extend the popular BERT model \cite{devlin2018bert} by adding masked multi-head attention to capture the graphical structure of ASTs. We evaluate the impact of this addition in \sect{sec:baselines}.


{\modelname} feeds the input code and natural language representations to an encoder, which is composed of a stack of $N = 4$ identical layers (\fig{fig:coral}).
Similar to Transformers \cite{vaswani2017attention}, we equip each layer with a multi-head self-attention sublayer and a feed-forward sublayer. 
The graph structure is captured through \emph{masked} attention (\equationref{aggregate_formula} below).

\xhdr{Masked Multi-Head Attention}
We use $Aggregate_k^i$ to represent the self-attention function of $head_i$ in $layer_k$. 
Let $(q,k,v)$ be the query, key, and value decomposition of the input to $Aggregate_k^i$. Queries and keys are vectors of dimension $d_k$, and values are vectors of dimension $d_v$. For a given node $u$, let $(q_u, k_u, v_u)$ be the triple of query, key and value, and let $N(u)$ be the set of all its neighbours. Formally, the parameters $q_u, k_u, v_u$ vary across each $head_i$ and $layer_k$, but we drop additional notation for simplicity here. Then we compute aggregate results as: 
\vspace{-5pt}
\begin{equation}\label{aggregate}
Aggregate_k^i(u)=\Sigma_{v\in N(u)} Softmax(\frac{q_u\cdot k_v}{\sqrt{d_k}})\cdot v_u
\end{equation}


We adopt the scaling factor $\frac{1}{\sqrt{d_k}}$ from Vaswani \etal \cite{vaswani2017attention} to mitigate the the dot product's growth in magnitude with $d_k$. 
In practice, the queries, keys, and values are assembled into matrices $Q, K, V$. We compute the output in matrix form as: 
\vspace{-5pt}
\begin{equation}\label{aggregate_formula}
Aggregate_k^i(Q,K,V)=Softmax(\frac{\tilde{A}\odot QK^T}{\sqrt{d_k}})V
\end{equation}
\noindent where $\tilde{A}=A+I$ is the adjacency matrix with self-loops added to implement the masked attention approach, where each node only attends to its neighbours (described in \sect{subsec:input_graphs}) and itself.

Since we adopt multi-head attention, we concatenate $h$ heads within the same layer:
\begin{equation}
    MultiHead(Q,K,V) = Concat(head_1,...,head_h)W_O 
\end{equation}
\begin{equation}
    head_i=Aggregate_k^i(XW^i_Q, XW^i_K, XW^i_V)
\end{equation}
\noindent where $head_i\in\mathbb{R}^{d_v}$ and $W^i_Q\in\mathbb{R}^{d_{model}\times d_k}$, $W^i_K\in\mathbb{R}^{d_{model}\times d_k}$, $W^i_V\in\mathbb{R}^{d_{model}\times d_v}$, and $W_O\in\mathbb{R}^{h*d_v\times d_{model}}$ are projection matrices that map the node embeddings $X$ to queries, keys, values, and multi-head output, respectively.

\xhdr{Feed Forward}
In each layer, we additionally apply a fully connected feed-forward sublayer. This is composed of two linear transformations with ReLU activation in between:
\begin{equation}
    FFN(x) = W_{FF2}\cdot max(0,W_{FF1}\cdot x+b_{FF1})+b_{FF2}
\end{equation}
where $W_{FF1}\in \mathbb{R}^{h*d_{model}\times d_{model}}$, $W_{FF2}\in\mathbb{R}^{d_{model}\times h*d_{model}}$, $b_{FF1}$ and $b_{FF2}$ are parameters learned in model. 

\xhdr{Add \& Norm}
Each sublayer is followed by layer normalization~\cite{ba2016layer}. The output of each sublayer is:
\begin{equation}
LayerNorm(x + Sublayer(x))
\end{equation}
where  $Sublayer(x)$ is multi-head attention or feed forward.

\xhdr{Output}
The multi-head attention sublayer and feed-forward sublayer are stacked and make up one ``layer''. 
After stacking this layer four times, the encoder's output contains representations of all the nodes in the input sequence. 
We take the embedding of the [CLS] node as the representation of the  each notebook cell's graph (Section~\ref{subsec:input_graphs}), denoted as $z\in\mathbb{R}^{d_{model}}$. 

We compress this cell representation $z$ into a lower-di\-mens-ion\-al  distribution over $K$ ``topics'' to capture information about the data analysis stages. 
Concretely: 
\begin{equation}
\label{eq:topic}
p_{topic} = Softmax(W_{topic}\cdot z+b)
\end{equation}
where $W_{topic}\in  \mathbb{R}^{ K\times d_{model}}$ is the weighted matrix parameter and $b$ is the bias vector. 

\subsection{Weak Supervision}
\label{sec:weak_supervision}

It is prohibitively expensive to obtain manual annotations of data analysis stages at scale, as doing so would require thousands of  person-hours of work by domain experts. Therefore, we  use five simple heuristics to tailor \modelname\ to the prediction task described in \sect{sec:task_description}:

\begin{enumerate}[leftmargin=*]
\item We collect a set of seed functions and assign each to a corresponding stage based on its usage. Any cell that uses a seed is weakly labeled as the corresponding stage. For example, any cell that calls "sklearn.linear\_model.LinearRegression" is weakly labeled \MODEL.
The full set of 39 seed functions is available in Appendix \ref{app:seeds}. 
We demonstrate \modelname's ability to correctly classify \emph{unseen} code outside these  functions in \sect{sec:error_analysis}. 
\item A cell with one line of code that does not create a new variable is weakly labeled \EXPLORE. This rule leverages a common pattern in Jupyter notebooks where users often use single line expressions to examine a variable, such as a DataFrame.
\item A cell with more than 30\% import statements is labeled \IMPORT.
\item A cell whose corresponding markdown is less than four words and contains \{‘logistic regression’, ‘machine learning’, ‘random forest’\} is weakly labeled \MODEL.
\item  A cell whose corresponding markdown is less than four words and contains ‘cross validation’ is weakly labeled \EVALUATE.
\end{enumerate}
Note that there may be conflicts between these rules.
We observe that less than one percent of cells in our corpus comply with more than one of these heuristics, further supporting our decision to formulate labels as mutually exclusive. We resolve any such conflicts by assigning priority in the following order: \IMPORT, \MODEL, \EVALUATE, \EXPLORE, \WRANGLE.\footnote{We also include a dummy sixth stage to represent cells that are empty or not covered by one of these heuristics. To reflect the uncertainty of these stages they are not included in the model's loss function.}
In this layer, we aim to compute $p_{stage}$ - a probability distribution over these six stages - from the topic distribution computed in \equationref{eq:topic}.
We implement this by mapping the topic distribution $p_{topic}$ to a probability distribution $p_{stage}$ over the $n_{stages}=6$ stages. We compute the stage distribution $p_{stage}$ as follows, where $W_{stage}\in \mathbb{R}^{K\times n_{stages}}$:
\begin{equation}\label{eq:model}
p_{stage}=softmax(W_{stage}\cdot p_{topic} + b_{stage})
\end{equation}



We adopt cross entropy loss to minimize classification error on weak labels.
For each $p_{topic}$, loss is computed as:
\begin{equation}\label{eq:cross_entropy_loss}
    L_{weakly\_supervised} = -\Sigma_{s}y_{o,s} log(p_s)
\end{equation}
where $y_{o,s}$ is a binary indicator (0 or 1) if stage label $s$ is the correct classification for observation $o$ and $p_s$ is the predicted probability $p_{stage}$ is of stage $s$. 

The five weak supervision heuristics cover about 20\% of notebook cells in the training data. 
To minimize the model's ambiguity on the remaining 80\% of unlabeled data, and encourage it to choose a stage for each topic, we add an additional loss function. Concretely, we add an entropy term to $p_{stage}$ to encourage uniqueness by forcing the topic distribution to map to as few stages as possible:
\begin{equation}\label{eq:entropy_loss}
    L_{unique\_stage} = -\Sigma_s p_s log (p_s)
\end{equation}
where $p_s$ is the predicted probability $p_{stage}[s]$ for stage $s$.
This entropy objective is minimized when $p_s = 1$ for some $s$ and $p_{s'} = 0$ all other $s'$.

\subsection{Unsupervised Learning Through Reconstruction}
\label{subsec:unsupervised_topic_model}

As the weak supervision heuristics only cover about 20\% of the cells, we enrich the model with additional training through an unsupervised topic model.
Here, the goal is to optimize the topic representation $p_{topic}$ such that we can reconstruct the intermediate cell representation $z$. 
We reconstruct $z$ from a linear combination of its topic embeddings $p_{topic}$:
\begin{equation}\label{reconstruction}
r=R\cdot p_{topic}
\end{equation}
where $R \in \mathbb{R}^{d_{model}\times K}$ is the learned cell embedding reconstruction matrix. 
This unsupervised topic model is trained to minimize the reconstruction error. We adopt the contrastive max-margin objective function using a Hinge loss formulation~\cite{weston2011wsabie,socher2014grounded,iyyer2016feuding}.
Thus, in the training process, for each cell, we randomly sample $m = 5$ cells from our dataset as negative samples:
\vspace{-5pt}
\begin{equation}\label{hinge_loss}
L_{unsupervised}=\Sigma_{c\in D}\Sigma_{i=1}^{m=5} \text{max}(0,1-r_c z_c+r_c n_i)
\end{equation}
where D is the training data set, $r_c$ is reconstructed vector of cell $c$, $z_c$ is intermediate representation of cell $c$, and $n_i$ is the reconstructed vector of each negative sample. This objective function seeks to minimize the inner product between $r_c$ and $n_i$, while simultaneously maximizing the inner product between $r_c$ and $z_c$.

We also employ a regularization term from He \etal \cite{he2017unsupervised} to promote the uniqueness of each topic embedding in $T$:
\begin{equation}\label{unsupervised_unique}
L_{unique\_topic}=\left\|R_{norm}\cdot R_{norm}^T-I\right\|
\end{equation}
where $I$ is the identity matrix and $R_{norm}$ is the result of L2-row-normalization of $R$. This objective function reaches its minimum when the inner product of two topic embeddings is 0. 
We demonstrate in \sect{sec:ablation} 
that this additional unsupervised training improves overall classification performance.
%

\subsection{Final Optimization Objective}
\label{subsec:optimization_objective}
We combine the loss functions of Equations~\eqref{eq:cross_entropy_loss},\eqref{eq:entropy_loss},\eqref{hinge_loss}, and \eqref{unsupervised_unique}\ 
into the final optimization objective:
\begin{equation}
\label{eq:loss}
\begin{aligned}
    L =& \; \lambda_{1}L_{weakly\_supervised} + \lambda_{2}L_{unique\_stage}\\
         &+\lambda_{3}L_{unsupervised} + \lambda_{4}L_{unique\_topic}
\end{aligned}
\end{equation}
where $\lambda_{1} $, $\lambda_{2}$, $\lambda_{3}$ and $\lambda_{4}$ are hyperparameters that control the weights of optimization objectives.  


We experiment with various training curricula and find that CORAL with the hyperparameters in described in Appendix~\ref{app:experiment_setting} achieves the best loss (\equationref{eq:loss}) on the validation set.
Importantly, this optimization and model training is based on solely on the labels from weak supervision heuristics.
We do not use expert annotations (\sect{sec:expert_annotated_notebooks}), 
which we exclusively reserve for the final evaluation.
\section{Evaluation}
\label{sec:evaluation}
{\modelname} achieves accuracy of over 72\% on the stage classification task using an unseen test set (Section~\ref{sec:expert_annotated_notebooks}), outperforming a range of baseline models and demonstrating that weak supervision, unsupervised topic modeling, and adding markdown information all strictly improve overall classification performance.


\figBaseline

\subsection{Baseline Comparison}
\label{subsec:baseline_comparison}
\label{sec:baselines}
In Figure \ref{fig:performance_with_baselines}(a) we compare {\modelname}'s performance to eight baselines, which we describe below. Importantly, the lack of ground truth labels in our training set makes it impossible to evaluate a model that does not use some amount of weak supervision, as without these heuristics we cannot map between learned topics and data science stages.

\xhdr{Expert Heuristics (Weak Supervision Only)}
How well does a simple baseline perform that considers only library information? 
For example, \textit{pandas} is commonly used to wrangle data, and \textit{scikit-learn} is common in modeling. 
We compare against an improved version of this baseline, where we include \emph{all} expert heuristics described in \sect{sec:weak_supervision}. This set of heuristics consider function-level and markdown information in addition to library information. This is a natural comparison since this is the exact weak supervision used in \modelname. These heuristics cover only 20.38\% of the test examples, so we choose one stage uniformly at random otherwise. 

\xhdr{LDA Representation + Weak Supervision}
How important is it to use a deep neural encoder for our task?
To address this question, we replace {\modelname}'s encoder with a Latent Dirichlet Allocation (LDA)~\cite{blei2003latent} topic model, but use the same input data (\sect{subsec:input_graphs}), and the same weak supervision (Section \ref{sec:weak_supervision}). 
Specifically, we optimize this model with $L_{weak\_supervision}$ (Eq.~\eqref{eq:cross_entropy_loss}) and $L_{unique\_stage}$ (Eq.~\eqref{eq:entropy_loss}) on top of the unsupervised LDA representation.
We first used the same number of LDA topics (50) as we use in \modelname ~(i.e. the size of the cell representation $p_{topic}$). However, this baseline only performed at the level of the Expert Heuristics Only baseline. In order to make this baseline stronger we doubled the number of LDA topics to 100, which did improve performance.

\xhdr{Neural Baselines}
How well does a noncontextual neural model perform on our task? What are the benefits of using the graphical structure of ASTs instead of treating code cells as sequences of tokens in deep neural networks? How important is the multitask objective that combines weak supervision heuristics and an unsupervised topic model?  To address these questions, we compare \modelname\ against the noncontextual Word2Vec~\cite{mikolov2013distributed} model and the state-of-the-art language model, BERT~\cite{devlin2018bert}, which have both been previously applied to source code representation learning \cite{feng2020codebert,kanade2019pretrained}. We trained all neural baselines with both markdown and code using the same pre-training corpus as \modelname{}.  To explore the sensitivity of these models to their input representations, we tried both treating the code as sequences of tokens and as serialized ASTs. For the BERT baselines we used the standard architecture with the same embedding size as \modelname\ and masked language model pretraining. Predictions are made with a single layer using the same weak supervision heuristics as \modelname{}. We evaluated BERT baselines both with and without ASTs and finetuning. When finetuning, we backpropogated the single layer's loss through the encoder. After pre-training, we optimized the model with $L_{weak\_supervision}$ (Eq.~\eqref{eq:cross_entropy_loss}) and $L_{unique\_stage}$ (Eq.~\eqref{eq:entropy_loss}) on top of the learned representations of code cells.

\xhdr{Results}
Results from these experiments are available in Figure~\ref{fig:performance_with_baselines}(a). The Expert Heuristics baseline achieves 34.1\% accuracy on the unseen expert annotations described in \sect{sec:expert_annotated_notebooks}. Even though it uses the same amount of supervision, {\modelname} is 38\% more accurate than this baseline, demonstrating that {\modelname} learns significantly more than simply memorizing the heuristic rules.
\modelname{}\ also favorably compares to state-of-the-art neural language models, beating the highest performing BERT baseline by 4.3\%. We observe that while popular deep learning techniques like finetuning produce only a marginal difference in model performance, \modelname\ significantly outperforms all other baselines (Wilcoxon signed rank, $p<0.001$).

\subsection{Ablation Study}
\label{sec:ablation}

\tableMaxSeq
\tableWeakSupervision
\tableTrainSize

We just demonstrated in \sect{subsec:baseline_comparison} that {\modelname} improves significantly over expert heuristics, representations that do not leverage graphical structure, and state-of-the-art neural models. Here we show that (1) adding markdown information, (2) weak supervision, and (3) additional unsupervised training all independently improve the performance of {\modelname}, as shown in Figure~\ref{fig:performance_with_baselines}(b).
Across all experiments we use maximum sequence length of $M=160$ and train on the maximum 1M code cells, based on the best performing model overall.

\xhdr{\modelname~without Markdown}
For this ablation, we remove any markdown information from the input sequence, while keeping all other aspects of {\modelname} the same. We compare maximum sequence length of 80, 120 and 160 since the maximum sequence length $M$ may interact with markdown information due to truncation (\sect{subsec:input_graphs}). 
We find that including markdown information consistently and significantly improves performance 12\%  at $M=160$, even though less that 9\% of cells are directly preceded by markdown  (Table~\ref{tab:max_seq_length}). Furthermore, these comparatively rare comments significantly improve performance even on cells that \textit{do not have corresponding markdown information} from 59.6\% to 72.6\%, suggesting that markdown cells help \modelname\ better represent source code independent of these comments. 




\xhdr{\modelname~with Less Weak Supervision}
The weak supervision heuristics described in \sect{sec:weak_supervision} 
cover about 20\% of the training examples.
We simulate lower coverage by randomly subsampling  50\% and 25\% of these weakly labeled examples (\ie 10\% and 5\% of all examples). 
Higher weak supervision coverage dramatically increases performance, but even at 25\% of examples {\modelname} still outperforms \modelname\ (No Masked Attention) by 10\% and BERT by 15\% (Table ~\ref{tab:weak_super}).




\xhdr{\modelname~without Unsupervised Topic Model}
This baseline evaluates the marginal benefit of {\modelname}'s unsupervised topic model. Specifically, we remove  $L_{unsupervised}$ (Eq.~\ref{hinge_loss}), and $L_{unique\_topic}$ (Eq.~\ref{unsupervised_unique}) from {\modelname} but keep everything else the same.
We show that the unsupervised training objective improves overall accuracy by 
10\% (Figure~\ref{fig:performance_with_baselines}(b)). This demonstrates the significant potential of combining limited weak supervision with additional unsupervised training in a multi-task framework. 




\figErrorAnalysis
\tableHoldOutSeeds

\subsection{Impact of Input Length \& Training Set Size}
\label{sec:additional_experiments}

\xhdr{Maximum Sequence Length}
We investigate how model performance changes with the maximum input sequence length $M$ (see Table~\ref{tab:max_seq_length}). For {\modelname} models with and without markdown, a larger maximum sequence length consistently improves accuracy.  Longer sequence lengths may include more markdown information and limit truncation of larger cells. 
Only 6\% of the training examples have more than 160 nodes, and increases in $M$ also increase training time and memory requirements. Therefore, we did not consider models beyond $M=160$ and use this setting for all other experiments.

\xhdr{Training Dataset Size}
We evaluate the accuracy of {\modelname} and two other high-performing models with different training dataset sizes to gauge how sensitive our model is to training data size. We fix $M$ to 160 and train with a maximum of 1M notebook cells. In all other experiments, we use the maximum 1M notebook cells for training. 
While performance consistently decreases with smaller training data (Table~\ref{tab:train_size}), {\modelname} achieves an accuracy of 
61.85\%  with only 1k examples and outperforms baselines by a large margin. 
This demonstrates that the {\modelname} architecture is effective at learning useful code representations even in smaller-data scenarios, such as on the order of magnitude of a typical GitHub repository. 


\subsection{Error Analysis}

\label{sec:error_analysis}
\xhdr{Confusion Matrix}
We include a confusion matrix of {\modelname}'s predictions from the best performing model ($M=160$ trained on 1M examples) in Appendix \ref{app:confusion_mat} (Figure \ref{fig:model_confusion}).
The most frequent confusion is misclassifying \EXPLORE~as \WRANGLE. This is in part because \WRANGLE\ and \EXPLORE\ are the two most common stages in the hand labeled corpus, but also possibly because analysts may apply simple transformations while primarily using a cell to visualize or otherwise explore data.


\xhdr{Unseen Functions}
To evaluate how well {\modelname} can learn beyond memorizing examples from weak supervision, we select eight common data analysis function and compare the labels of cells that contain them (Table \ref{tab:hold_out_seeds}). Importantly, these functions were not used in weak supervision and thus were never directly associated with any label in the model. Many functions demonstrate clear stage membership in line with our expectations (\eg \textit{\seqsplit{pandas.DataFrame.groupby}}, \textit{seaborn.countplot}),
demonstrating that \modelname~can assign cells including these functions to likely correct stages.
Other functions exhibit a more even distribution across stages. For example, \textit{\seqsplit{sklearn.linear\_model.PassiveAggressiveClassifier}}, a simple linear classifier, appears in both {\MODEL} and {\EVALUATE} cells. While ambiguity between stages is rare overall (\sect{sec:expert_annotated_notebooks}) we hypothesize that this confusion may be the result of the scikit-learn use pattern where users specify and evaluate their models in the same cell.

\xhdr{Example Predictions}
We highlight three predictions in Figure \ref{fig:error_analysis} to demonstrate {\modelname}'s ability to capture data analysis semantics and inherent ambiguity. 
In Figure \ref{fig:error_analysis}(a), the user transforms a \textit{pandas} DataFrame and calls \textit{pandas.DataFrame.groupby}, a function typically used to aggregate data. While a naive method (\eg the expert heuristic baseline in \sect{sec:baselines}) might label the cell as \WRANGLE, {\modelname} infers that the analyst's intention is to use this user-defined function to evaluate a classifier with a confusion matrix, likely making use of the information in the comment and function parameters, and appropriately labels the cell as {\EVALUATE}.

In Figure \ref{fig:error_analysis}(b), the analyst loads data, selects a subset, creates a plot, and fits a linear regression. {\modelname} correctly identifies this example as serving to both modify data and look for patterns, but assigns a higher probability to {\EXPLORE}, demonstrating its ability to capture the significance of previously unseen statistical visualization methods like \textit{seaborn.regplot}. 

In Figure \ref{fig:error_analysis}(c), the analyst calls a user-defined function. While \modelname\ has never seen this function or notebook, it still correctly identifies the intent of the cell as \EXPLORE\, likely by attending to tokens like \say{plot} and \say{breakdown}.

\section{Large Scale Studies of Scientific Data Analysis}
\label{sec:large_scale_studies}

Our model and datasets provide an opportunity to pose and answer previously unaddressable questions about the data analysis process, the role of scientific analysis in academic publishing, and differences between scientific domains. We note that our corpus (\sect{sec:jupyer_notebooks}) is limited to the most recent (potentially partial) snapshot of the user's analysis and that the observational nature of this data prohibits any causal claims. 
\figFracCode

\subsection{Are There Differences Between Academic Notebooks and Non-Academic Notebooks?}
\label{sec:academic_vs_non_academic}
Differences between academic and non-academic notebooks could identify how practices vary across these communities. 

\xhdr{Method}
The Semantic Scholar Open Research Corpus (S2ORC) is a publicly available dataset containing 8.1M full-text academic articles \cite{lo-wang-2020-s2orc}. In order to relate these papers to relevant source code, we performed a regular expression search across the corpus for any reference to a GitHub repository, returning associations between 2.0k papers and 7.1k notebooks from the UCSD corpus. We use this dataset to resolve previously unanswerable questions about the role of analysis code in the scientific process. Although there is no strict guarantee that a linked notebook contains the data analysis that was used to create the paper, the median notebook is linked to exactly one paper, indicating some degree of injectivity from notebooks to papers. Furthermore, manual inspection of our dataset and prior work indicate that researchers often break their analysis up across many notebooks, which may explain why papers link to multiple notebooks. So as not to bias our analysis against how a scientist decides to structure their code, we compute statistics for each paper by concatenating all associated notebooks.
We compute the fraction of code devoted to each data analysis stage and the fraction of cells that are followed by a cell of a different stage and examine differences between academic and non-academic notebooks.

\xhdr{Results}
Academic notebooks devote 56\% more code to exploring data and 26\% less code to developing models than non-academic notebooks (Figure~\ref{fig:frac_code_per_stage}(a)). Furthermore, we note that analysts on average use only 23\% of their code for the traditionally boring and laborious process of wrangling data. While the relative size of the stage likely does not accurately reflect the relative \emph{effort} of data wrangling, it is perhaps surprising that such a maligned stage of the process~\cite{kandel_enterprise_2012} is represented by a comparatively low fraction of all code. 
We also find significant differences in the fraction of cells that are followed by a cell of a different stage (Figure~\ref{fig:frac_code_per_stage}(b)). Most interestingly, cells are in general more likely than not to transition to a different stage. This result supports the hypotheses that notebooks follow a transitory process through the data science process to complete an analysis rather than dwelling on any particular stage.\


\subsection{Is the Content of Notebooks Related to the Impact of Associated Publications?}
Evidence of a relationship between scientific notebooks and publication impact may encourage researchers to publish their code, and could reveal differences between the priorities placed on scientific data analysis by different domains.   
\label{sec:ref_assoc_content}

\xhdr{Method}
 We employ a negative binomial regression to estimate the impact of notebook stage distribution on the number of citations their associated papers receive. We hypothesize that notebooks which evenly and comprehensively document their analysis (rather than focusing on just one part) may receive more citations. In our first regression \texttt{R1}, we therefore regress citation count on the $\text{Stage Entropy} = -\sum_{k}{p_k}\log{p_k}$, where $p_k$ is the fraction of the notebook that is devoted to stage $k$. This captures the uniformity of the distribution of stages across a paper's associated notebooks. Here, we normalized this quantity across all publications by taking the Z-score. We controlled for a paper's year of publication and domain. To reveal differences between disciplines, we build upon this experiment with a second regression \texttt{R2}, which includes all terms from \texttt{R1} except for the entropy term, but adds interaction variables between the Z-scores of the fraction of each paper's notebook devoted to each data analysis stage and paper domains to capture differences between disciplines. additional details for these regression models are available in Appendix \ref{app:reg_details}.

\figPerDomain

\xhdr{Results} We find that papers that link to notebooks have \\ $10^{\beta_{hasNotebook}} = 10^{1.34} \approx 21.88$ times more citations than papers that do not reference a notebook (95\% CI: [1.29, 1.41], $p < 0.001$). From \texttt{R1} we note that Stage Entropy is strongly related to the number of citations a publication receives, as those publications can expect a $10^{\beta_{stageEntropyZ}} = 10^{0.33} \approx 2.11$  times increase in citations with an entropy level for each standard deviation above the mean (95\% CI: [0.26, 0.39], $p<0.001$) This result suggests that researchers may value notebooks which evenly document the whole data science process, rather than highlighting just one part of analysis. These results also indicate that a notebook with one standard deviation more than the average {\EXPLORE} code would expect $10^{\beta_{EXPLORE}} = 10^{-0.4325} \approx 0.35$ times the citations in its associated paper than a notebook with an average quantity of all stages (95\% CI: [-0.64,-0.22], $p<0.001$). One possible explanation for this effect is that notebooks which feature a high volume of code for exploring data are associated with generating hypotheses, and may therefore be associated with incomplete or exploratory publications that are less likely to attract references. 

The results from \texttt{R2} (Figure \ref{fig:per_domain_betas}) indicate significant differences between domains. Most notably, we find that in computer science and mathematics an increase in the portion of code devoted to wrangling data decreases the citation count in expectation, while no such interaction is present for papers from biological sciences. We hypothesize that the most popular cited notebooks in computer science and mathematics may cleanly demonstrate new techniques and models, rather than documenting an extensive data wrangling pipeline. 

We note that although these effect sizes may seem large, we need to consider that the median citation count for papers is only two. This implies that even with a high citation multiplier, papers with just a few citations would expect a rather moderate increase in citations. 

\section{Conclusion}
We presented {\modelname}, a novel weakly supervised neural architecture for generating representations of code snippets and classifying them as stages in the analysis pipeline. We showed that this model outperforms a suite of baselines on this new classification task. Further, we introduced and made public the largest dataset of code with associated publications for scientific data analysis, and employed {\modelname} to answer open questions about the data analysis process.



\bibliographystyle{IEEEtran}
\bibliography{references}

\clearpage
\onecolumn
\appendix

\begin{center}
   \textsc{Reproducability} 
\end{center}


\subsection{Weak Supervision Seed Functions}
\label{app:seeds}
The seed functions with associated data analysis stages used in weak supervision heuristics are listed in Table~\ref{tab:seed-functions}.

\tableSeed






\subsection{Experiment Setting}
\label{app:experiment_setting}
We train {\modelname} with 1M cells on a single GeForce RTX 2080 Ti GPU. 
The model has four attention heads and four layers of dimension $d_{model}$ = 256. We set the number of topics (\sect{subsec:encoder_attention}) to 50 and maximum sequence length ($M$) to 160. We set $\lambda_{1}=0.1$, $\lambda_{2}=0.3$, $\lambda_{3}=1$ and $\lambda_{4}=1$.
We train the model by minimizing $L$ in Equation~\eqref{eq:loss}, using the SGD optimizer with a learning rate $\alpha=1\times 10^{-5}$, $\beta=0.9$.
Training is done on mini-batches of size 16, for up to 8 epochs with an early stopping criteria if validation error had not improved for 3 epochs.
Each epoch takes about 2.5 hours to train. 

\newpage
\subsection{Algorithm}
\label{app:coral_algo}
The CORAL Algorithm is shown in Figure~\ref{fig:pseudo_code}.

\vspace{-10pt}
\figPseudoCode

\subsection{Qualitative Rubric}
\label{app:qualitative_rubric}
The qualitative rubric used for labeling the Expert Annotated Dataset (Section~\ref{sec:expert_annotated_notebooks}) used for final model evaluation is listed in Table~\ref{tab:rubric}.
\tableRubric

\subsection{Confusion Matrix}
\label{app:confusion_mat}
The confusion matrix for \modelname's predictions on the data analysis stage prediction task is shown in Figure~\ref{fig:model_confusion}.
\figConfusion

\subsection{Regression Details}
\label{app:reg_details}

The following details apply to both regression \texttt{(R1)} and regression \texttt{(R2)}. We chose to use a negative binomial for zero-inflated counts regression because we observed that the mean number of citations (8.52) was substantially less than the variance (1,308). We expect that a paper's year of publication will influence its citation count, and therefore we control for this variable. We also expect each paper's domain to be related to notebook characteristics, so we limit our analysis to the three most common domains in GORC 
and control for this factor using indicator variables. We note that our analysis does not substantially change with the inclusion of the top five, 10, or 20 domains. If a paper is linked to more than one notebook, for the purpose of these regressions, we concatenate the notebooks and calculate statistics across this concatenation.

\end{document}